\documentclass[10pt]{article}
\usepackage[a4paper,margin=0.82in]{geometry}
\usepackage{graphicx}
\usepackage{booktabs}
\usepackage{amsmath,amssymb,amsfonts,mathtools}
\usepackage{amsthm}
\usepackage{bm}
\usepackage{multirow}
\usepackage{microtype}
\usepackage{enumitem}
\usepackage[backend=biber,style=numeric-comp,sorting=none,natbib=true,url=false,doi=true,eprint=true]{biblatex}
\usepackage[colorlinks=true,allcolors=blue]{hyperref}
\newtheorem{lemma}{Lemma}
\newtheorem{proposition}[lemma]{Proposition}
\newtheorem{theorem}[lemma]{Theorem}

\setlist{nosep}
\title{Structured Proper Loss Geometries for Multiclass Classification:\\Theory and Controlled Empirical Evaluation}
\author{Soumyadip Sarkar\\\small Independent Researcher}
\date{}
\begin{document}
\maketitle
\begin{abstract}
Strictly proper scoring rules identify the true conditional class distribution at population level, but their curvature can alter optimization and finite-sample behavior. We study three multiclass objectives: a class-aware quadratic Bregman score (CAPM), a strongly convex generator with constrained log-cosh ridges (HPG), and an HPG objective with an annealed probability-margin penalty (APMS). CAPM is treated as a structured instance of established quadratic scoring-rule theory. We derive conditional-regret, curvature, range, and logit-gradient bounds for CAPM and HPG, and prove exact penalty-range and conditional-target displacement bounds for APMS. Controlled five-seed experiments use Digits, Wisconsin breast cancer, and synthetic confusion and long-tail problems under clean labels, symmetric and pair-flip corruption, class imbalance, calibration evaluation, input corruption, and first-order adversarial perturbations. The candidates are close to cross-entropy on clean data and show descriptive gains in some noisy-label cells, but the five-seed comparisons are interpreted descriptively rather than as significance evidence. The selected noisy-label baselines perform better on Digits with 40\% symmetric label noise, and explicit prior-adjustment methods perform better in the 30:1 synthetic long-tail experiment. Ablations do not show a consistent benefit from the candidate-specific graph, ridge, or margin components. The mathematical analysis establishes the stated properties, and the experiments delimit the empirical evidence; together they do not support a claim of general superiority.
\end{abstract}

\section{Introduction}\label{sec:introduction}

The loss function used to train a probabilistic classifier determines the population quantity being estimated and affects the geometry through which model parameters are optimized. This paper uses softmax cross-entropy as the reference baseline; the corresponding logarithmic score is strictly proper under the usual probability-level interpretation \citep{gneiting2007proper}. Nevertheless, finite neural networks trained with cross-entropy can be miscalibrated, and sufficiently expressive networks can fit corrupted labels \citep{guo2017calibration,arpit2017memorization}. These observations have motivated objectives designed for particular failure modes, including focal loss for class imbalance in dense detection \citep{lin2017focal}, generalized and symmetric cross-entropy for noisy labels \citep{zhang2018gce,wang2019sce}, and prior- or margin-adjusted objectives for long-tailed recognition \citep{cui2019classbalanced,cao2019ldam,ren2020balanced,menon2021logit}.

Strictly proper scoring rules provide a statistical constraint: at the population level, their conditional risk is uniquely minimized by the true class-probability vector \citep{gneiting2007proper,ovcharov2015proper}. Properness does not, however, determine finite-sample accuracy, calibration after model misspecification, optimization speed, robustness to corrupted labels, or performance under class-prior shift. The curvature of a proper score can therefore matter even when different scores share the same population minimizer. Existing work on composite losses, learned proper losses, and task-tailored proper scores shows that proper objectives can have adaptable geometry \citep{reid2010composite,reid2012multiclass,lam2023legendretron,plaud2026tailoring}.

This paper studies three multiclass objectives. The first, class-aware proper Mahalanobis loss (CAPM), is a structured quadratic Bregman score. The second, hyperbolic proper generator loss (HPG), adds constrained log-cosh ridge terms to a quadratic generator. The third, annealed probability-margin shaping (APMS), augments HPG with a bounded margin penalty whose coefficient is scheduled toward zero. CAPM is an application-specific parameterization of classical quadratic proper-score theory rather than a new scoring-rule family. HPG and APMS are analyzed as concrete constructions, with the contribution centered on their exact properties and controlled comparison rather than on a broad claim about adaptable proper geometry.

The study makes three contributions. First, it gives complete definitions and derives conditional regret, curvature, range, and gradient bounds for CAPM and HPG. Second, it proves the exact range of the APMS margin penalty on the simplex and establishes both square-root and linear bounds on the displacement of its conditional minimizer from the true probability vector. Third, it reports cross-entropy comparisons across the evaluated regimes and targeted comparisons with established noisy-label, long-tail, and calibration baselines. The empirical results are deliberately interpreted at the scale supported by the experiments: the proposed structures are close to or slightly above cross-entropy in several cells but do not consistently outperform specialized noisy-label or long-tail methods.

\section{Related work}\label{sec:related}

\subsection{Proper scoring rules and composite losses}
The Brier score is a classical quadratic probability score \citep{brier1950verification}. General proper-scoring-rule theory characterizes losses whose expected value is optimized by truthful probabilistic reporting \citep{gneiting2007proper}. For differentiable finite-outcome scores, convex entropies and Bregman divergences provide a standard representation under appropriate regularity assumptions \citep{ovcharov2015proper}. Composite-loss theory separates a probability-level proper loss from a link that maps model outputs to probabilities and characterizes convexity and classification calibration \citep{bartlett2006convexity,reid2010composite,reid2012multiclass}. Recent studies have trained language models with non-logarithmic proper scores \citep{shao2024language}, learned proper multiclass losses \citep{lam2023legendretron}, and tailored proper-score curvature to downstream estimation error \citep{plaud2026tailoring}. These works motivate treating the present contribution as a specific construction and evaluation rather than as a broad novelty claim based solely on replacing log loss with a parameterized proper geometry.

\subsection{Learning with noisy labels}
Robustness to label noise is model- and assumption-dependent. Symmetric-loss analyses establish risk invariance under idealized corruption models \citep{ghosh2017robust}. Loss-correction methods use an estimated transition matrix \citep{patrini2017loss}. Generalized cross-entropy interpolates between cross-entropy and an MAE-like objective \citep{zhang2018gce}; symmetric cross-entropy combines ordinary and reverse cross-entropy \citep{wang2019sce}; bi-tempered logistic loss modifies logarithmic and exponential functions in a Bregman construction \citep{amid2019bitempered}; and normalized active--passive losses address the underfitting that can occur with bounded or symmetric objectives \citep{ma2020normalized}. Synthetic symmetric and pair-flip noise are useful controlled settings but do not reproduce the heterogeneity of human annotation errors represented by datasets such as CIFAR-N \citep{wei2021cifarn}.

\subsection{Long-tailed learning, calibration, and adversarial evaluation}
Long-tailed learning methods alter sample weights, margins, or class priors. Class-balanced loss uses the effective number of samples \citep{cui2019classbalanced}; LDAM introduces class-dependent margins \citep{cao2019ldam}; Balanced Softmax modifies the softmax normalization using class frequencies \citep{ren2020balanced}; and logit adjustment incorporates empirical priors during or after training \citep{menon2021logit}. Focal loss with a positive focusing parameter is classification calibrated but is not a proper probability loss without a correction map \citep{charoenphakdee2021focal,komisarenko2024focal}. Temperature scaling is a simple post-hoc calibration method that preserves the predicted class for positive temperatures \citep{guo2017calibration}.

Adversarial accuracy is distinct from probability-space curvature. FGSM and projected-gradient attacks are empirical evaluation methods \citep{goodfellow2015explaining,madry2018towards}; adversarial training objectives such as TRADES optimize an explicit robustness--accuracy trade-off \citep{zhang2019trades}; and attack ensembles such as AutoAttack are designed to reduce evaluation pitfalls \citep{croce2020autoattack}. Consequently, the gradient bounds derived below are not adversarial certificates.

\section{Structured loss formulations}\label{sec:methods}

\subsection{Preliminaries}
Let $Y\in\{1,\ldots,K\}$, let $\eta\in\Delta^{K-1}$ denote the conditional class distribution, and let $p\in\Delta^{K-1}$ be a reported distribution. For a differentiable convex function $F$ defined on a neighborhood of the simplex, its Bregman divergence is
\begin{equation}
D_F(q,p)=F(q)-F(p)-\langle\nabla F(p),q-p\rangle.
\label{eq:bregman}
\end{equation}
We use the negatively oriented categorical score
\begin{equation}
\ell_F(p,y)=D_F(e_y,p),
\label{eq:score}
\end{equation}
where $e_y$ is the $y$th standard basis vector. The following standard identity fixes the statistical interpretation of the constructions \citep{gneiting2007proper,ovcharov2015proper}.

\begin{lemma}[Bregman conditional regret]\label{lem:regret}
For $R_F(\eta,p)=\mathbb E_{Y\sim\eta}[\ell_F(p,Y)]$,
\begin{equation}
R_F(\eta,p)-R_F(\eta,\eta)=D_F(\eta,p).
\label{eq:conditional-regret}
\end{equation}
If $F$ is strictly convex, $\ell_F$ is strictly proper. If $F$ is $m$-strongly convex and $M$-smooth in Euclidean norm, then
\begin{equation}
\frac{m}{2}\|\eta-p\|_2^2\le D_F(\eta,p)\le
\frac{M}{2}\|\eta-p\|_2^2.
\label{eq:bregman-quadratic}
\end{equation}
\end{lemma}

Differentiation of Eq.~\eqref{eq:score} gives
\begin{equation}
\nabla_p\ell_F(p,y)=H_F(p)(p-e_y),
\label{eq:prob-gradient}
\end{equation}
where $H_F=\nabla^2F$. If $H_F(p)\preceq MI$, then $\|\nabla_p\ell_F\|_2\le M\sqrt{2}$. For $p=\operatorname{softmax}(z/T)$, the softmax Jacobian has operator norm at most $1/(2T)$, yielding
\begin{equation}
\|\nabla_z\ell_F\|_2\le \frac{M}{\sqrt{2}T}.
\label{eq:logit-gradient}
\end{equation}
This is a bound on the loss gradient with respect to logits, not with respect to the input.

\subsection{Class-aware proper Mahalanobis loss}
Let $u=K^{-1}\mathbf 1$ and define
\begin{align}
F_{\mathrm C}(p)&=\frac12(p-u)^\top A(p-u),\label{eq:capm-generator}\\
A&=\lambda I+BB^\top+\gamma L_G+\delta D_{\mathrm{tail}},
\label{eq:capm-matrix}
\end{align}
where $\lambda>0$, $B$ is any real matrix with $K$ rows, $L_G\succeq0$ is a graph Laplacian, $D_{\mathrm{tail}}\succeq0$ is diagonal, and $\gamma,\delta\ge0$. The resulting loss is
\begin{equation}
\ell_{\mathrm C}(p,y)=\frac12(e_y-p)^\top A(e_y-p).
\label{eq:capm-loss}
\end{equation}
The experiments construct $L_G$ from training-set class-centroid similarities and $D_{\mathrm{tail}}$ from training class counts. These quantities are descriptive structures; they are not assumed to recover a causal or semantic class graph.

\begin{proposition}[CAPM properties]\label{prop:capm}
Let $m=\lambda_{\min}(A)>0$ and $M=\lambda_{\max}(A)$. CAPM is strictly proper and
\begin{equation}
R_{\mathrm C}(\eta,p)-R_{\mathrm C}(\eta,\eta)
=\frac12(\eta-p)^\top A(\eta-p).
\end{equation}
Moreover, Eq.~\eqref{eq:bregman-quadratic} holds, $0\le\ell_{\mathrm C}(p,y)\le M$, and Eq.~\eqref{eq:logit-gradient} holds with the same $M$.
\end{proposition}

\subsection{Hyperbolic proper generator loss}
HPG uses the generator
\begin{equation}
F_{\mathrm H}(p)=s\left[
\frac{\lambda}{2}\|p-u\|_2^2+
\sum_{r=1}^{R}a_r\rho_r^2
\log\cosh\!\left(\frac{w_r^\top p-b_r}{\rho_r}\right)
\right],
\label{eq:hpg-generator}
\end{equation}
with $\lambda>0$, $a_r\ge0$, $\rho_r\ge\rho_{\min}>0$, $\|w_r\|_2\le1$, and $s>0$. Its gradient and Hessian are
\begin{align}
\nabla F_{\mathrm H}(p)
&=s\left[\lambda(p-u)+\sum_r a_r\rho_r\tanh(v_r)w_r\right],\label{eq:hpg-gradient}\\
H_{F_{\mathrm H}}(p)
&=s\left[\lambda I+\sum_r a_r\operatorname{sech}^2(v_r)w_rw_r^\top\right],
\label{eq:hpg-hessian}
\end{align}
where $v_r=(w_r^\top p-b_r)/\rho_r$. The HPG loss is $\ell_{\mathrm H}(p,y)=D_{F_{\mathrm H}}(e_y,p)$.

\begin{proposition}[HPG curvature and range]\label{prop:hpg}
Define
\begin{equation}
m=s\lambda,\qquad
M=s\left(\lambda+\sum_r a_r\|w_r\|_2^2\right).
\end{equation}
Then $mI\preceq H_{F_{\mathrm H}}(p)\preceq MI$ for all $p$. Therefore HPG is strictly proper, satisfies Eq.~\eqref{eq:bregman-quadratic}, obeys $0\le\ell_{\mathrm H}(p,y)\le M$, and satisfies Eq.~\eqref{eq:logit-gradient} with this value of $M$. Its Hessian is globally Lipschitz with the valid bound
\begin{equation}
L_H\le s\sum_r\frac{4a_r}{3\sqrt3\,\rho_r}\|w_r\|_2^3.
\label{eq:hessian-lipschitz}
\end{equation}
\end{proposition}

\subsection{Annealed probability-margin shaping}
For the HPG core, define the smoothed probability margin
\begin{equation}
m_\tau(p,y)=p_y-\tau\log\sum_{j\ne y}\exp(p_j/\tau)
\end{equation}
and the bounded penalty
\begin{equation}
r(p,y)=\nu\,\operatorname{softplus}\!\left(
\frac{\kappa-m_\tau(p,y)}{\nu}\right),
\label{eq:apms-penalty}
\end{equation}
where $\tau,\nu>0$. At optimization step $t$, with $\beta_0\ge0$, $T_s>0$, and $q>0$, APMS uses
\begin{equation}
\ell_{\mathrm A,t}(p,y)=\ell_{\mathrm H}(p,y)+\beta_t r(p,y),
\qquad
\beta_t=\beta_0\left(1-\min\{t/T_s,1\}\right)^q.
\label{eq:apms-loss}
\end{equation}
For $\beta_t>0$, the objective is not generally proper. The schedule is intended to make the terminal objective coincide with HPG, although an early-stopped checkpoint can be selected before $\beta_t$ reaches zero.

\begin{theorem}[APMS penalty range and target displacement]\label{thm:apms}
For $K\ge2$, the exact minimum and maximum of $m_\tau(p,y)$ over the simplex and labels are
\begin{equation}
m_{\min}=-\tau\log\{\exp(1/\tau)+K-2\},
\qquad
m_{\max}=1-\tau\log(K-1).
\label{eq:margin-minimum}
\end{equation}
Consequently,
\begin{equation}
\nu\,\operatorname{softplus}\!\left(\frac{\kappa-m_{\max}}{\nu}\right)
\le r(p,y)\le C_r:=\nu\,\operatorname{softplus}\!\left(
\frac{\kappa-m_{\min}}{\nu}\right).
\label{eq:penalty-bound}
\end{equation}
Let $R_0(\eta,p)$ be the conditional risk of a twice differentiable proper Bregman score generated by an $F$ satisfying $H_F\succeq mI$ and let
\begin{equation}
p_\beta^\star\in\arg\min_{p\in\Delta^{K-1}}
\{R_0(\eta,p)+\beta\,\mathbb E_{Y\sim\eta}[r(p,Y)]\}.
\end{equation}
Then
\begin{align}
\|p_\beta^\star-\eta\|_2
&\le\sqrt{\frac{2\beta C_r}{m}},\label{eq:apms-sqrt-bound}\\
\|p_\beta^\star-\eta\|_2
&\le\frac{\sqrt2\,\beta}{m}.
\label{eq:apms-linear-bound}
\end{align}
Thus the conditional minimizer converges to $\eta$ as $\beta\to0$, but positive-$\beta$ properness does not follow.
\end{theorem}

\section{Experimental design}\label{sec:experiments}

\subsection{Datasets and training regimes}
The experiments use two public scikit-learn datasets and two synthetic classification problems generated with fixed source seeds \citep{pedregosa2011scikit}. Table~\ref{tab:datasets} summarizes the design. For Digits, breast cancer, and synthetic confusion, each random seed defines stratified 60/20/20 training, validation, and test partitions. Feature standardization is fitted on the training partition only. Symmetric corruption replaces a selected training label uniformly with a different class; pair-flip corruption maps a selected label to its successor modulo $K$. Validation and test labels remain clean.

The long-tail problem begins with a balanced 10-class synthetic population. Only the training pool is exponentially subsampled, producing an observed largest-to-smallest class-count ratio of 30:1 for every retained seed. Validation and test splits remain approximately balanced. This design isolates training-prior imbalance but is not a substitute for a natural long-tailed benchmark.

\begin{table}[t]
\caption{Datasets and evaluated training regimes.}
\label{tab:datasets}
\centering\scriptsize
\setlength{\tabcolsep}{3pt}
\begin{tabular}{lrrl}
\toprule
Dataset & Samples & Classes & Training regimes\\
\midrule
Breast cancer & 569 & 2 & clean; symmetric 20\%, 40\%\\
Digits & 1,797 & 10 & clean; symmetric 20\%, 40\%; pair-flip 40\%\\
Synthetic confusion & 2,400 & 6 & clean; symmetric 20\%; pair-flip 40\%\\
Synthetic long-tail & 6,000 source & 10 & clean labels; 30:1 training imbalance\\
\bottomrule
\end{tabular}
\end{table}

\subsection{Models, optimization, and baselines}
All primary comparisons use a two-hidden-layer multilayer perceptron with ReLU activations and dropout 0.1. Hidden width is 96 for input dimension at least 24 and 64 otherwise. AdamW uses weight decay $10^{-4}$, gradient clipping at 5, a maximum of 45 epochs, a minimum of 12 epochs, and patience 8. The learning rate is fixed at $10^{-3}$ in the primary comparison. A separate sensitivity analysis evaluates $3\times10^{-4}$, $10^{-3}$, and $3\times10^{-3}$ using validation NLL. Early stopping selects the checkpoint with the lowest clean validation NLL; model and loss state are restored together.

CAPM and HPG are positively rescaled to match a common mean-curvature target. The geometry seed is fixed independently of the data and model seed, and all losses receive the same deterministic minibatch order within a dataset--regime--seed cell. CAPM class structure is computed from the training partition only.

The general baselines are cross-entropy, label smoothing \citep{szegedy2016rethinking}, Brier loss, categorical MAE, focal loss, generalized cross-entropy, symmetric cross-entropy, active--passive loss, Poly-1 \citep{leng2022polyloss}, and bi-tempered logistic loss. The long-tail experiment additionally includes class-balanced cross-entropy, Balanced Softmax, training-time logit adjustment, and LDAM. Hyperparameters are fixed before test evaluation, so the reported comparisons are claims about this retained experimental configuration rather than about exhaustive per-loss tuning.

\subsection{Metrics and statistical analysis}
The reported metrics are accuracy, balanced accuracy for the long-tail balanced test split, negative log-likelihood (NLL), multiclass Brier score, and 15-bin expected calibration error (ECE). ECE is interpreted jointly with accuracy and proper scores because a low-confidence, low-accuracy classifier can have a small binning-based calibration error. Temperature scaling is fitted on validation logits and evaluated separately on the test split \citep{guo2017calibration}.

Each reported cell uses five matched random seeds. Means and sample standard deviations summarize variability. Paired two-sided Wilcoxon signed-rank tests compare each loss with cross-entropy on matched seeds, followed by Holm correction \citep{wilcoxon1945individual,holm1979simple}. With five nonzero pairs, the smallest attainable exact two-sided Wilcoxon $p$-value is $0.0625$; the study is therefore descriptive and underpowered for conventional significance claims.

\subsection{Input-corruption and adversarial probes}
Models trained on clean Digits are evaluated under Gaussian noise ($\sigma\in\{0.1,0.2\}$), salt-and-pepper corruption (fraction 0.1), FGSM ($\epsilon\in\{0.05,0.1,0.2\}$), and projected-gradient attack ($\epsilon\in\{0.1,0.2\}$). Pixel intensities are scaled to $[0,1]$. The attacks maximize cross-entropy against each evaluated model. These tests measure local sensitivity of the trained models; they are not robustness certificates and do not replace adversarial training or standardized attack suites.

\section{Results}\label{sec:results}

\subsection{Overall predictive performance}
\begin{table*}[t]
\caption{Accuracy for cross-entropy and the three studied objectives. The synthetic long-tail row reports balanced accuracy on the balanced test split; all other rows report ordinary test accuracy. Values are mean $\pm$ sample standard deviation over five matched seeds. The table is descriptive and is not used to assert statistical significance.}
\label{tab:candidate-accuracy}
\centering\small
\resizebox{\linewidth}{!}{%
\begin{tabular}{llcccc}
\toprule
Dataset & Training regime & CE & CAPM & HPG & APMS\\
\midrule
Breast cancer & Clean & 0.975 $\pm$ 0.013 & 0.977 $\pm$ 0.013 & 0.977 $\pm$ 0.013 & 0.977 $\pm$ 0.013\\
Breast cancer & Symmetric 20\% & 0.968 $\pm$ 0.016 & 0.968 $\pm$ 0.016 & 0.970 $\pm$ 0.018 & 0.968 $\pm$ 0.017\\
Breast cancer & Symmetric 40\% & 0.875 $\pm$ 0.034 & 0.872 $\pm$ 0.024 & 0.872 $\pm$ 0.024 & 0.867 $\pm$ 0.032\\
Digits & Clean & 0.963 $\pm$ 0.010 & 0.967 $\pm$ 0.010 & 0.967 $\pm$ 0.012 & 0.967 $\pm$ 0.010\\
Digits & Symmetric 20\% & 0.939 $\pm$ 0.019 & 0.944 $\pm$ 0.012 & 0.948 $\pm$ 0.014 & 0.948 $\pm$ 0.016\\
Digits & Symmetric 40\% & 0.915 $\pm$ 0.021 & 0.921 $\pm$ 0.017 & 0.920 $\pm$ 0.018 & 0.921 $\pm$ 0.017\\
Digits & Pair-flip 40\% & 0.698 $\pm$ 0.038 & 0.734 $\pm$ 0.074 & 0.733 $\pm$ 0.054 & 0.732 $\pm$ 0.052\\
Synthetic confusion & Clean & 0.601 $\pm$ 0.019 & 0.602 $\pm$ 0.023 & 0.602 $\pm$ 0.025 & 0.603 $\pm$ 0.024\\
Synthetic confusion & Symmetric 20\% & 0.556 $\pm$ 0.044 & 0.563 $\pm$ 0.038 & 0.562 $\pm$ 0.036 & 0.561 $\pm$ 0.037\\
Synthetic confusion & Pair-flip 40\% & 0.423 $\pm$ 0.030 & 0.431 $\pm$ 0.036 & 0.432 $\pm$ 0.033 & 0.431 $\pm$ 0.037\\
Synthetic long-tail & 30:1 imbalance & 0.164 $\pm$ 0.029 & 0.163 $\pm$ 0.026 & 0.162 $\pm$ 0.028 & 0.162 $\pm$ 0.028\\
\bottomrule
\end{tabular}%
}
\end{table*}

Table~\ref{tab:candidate-accuracy} shows that CAPM, HPG, and APMS are close to cross-entropy on the clean breast-cancer, Digits, and synthetic-confusion tasks. Their largest descriptive accuracy differences relative to cross-entropy occur under pair-flip noise on Digits, where the candidate means are approximately 0.73 compared with 0.70 for cross-entropy. Variability is also larger in this cell. On the synthetic long-tail problem, all three candidates remain close to cross-entropy and have low balanced accuracy.

Because each cell uses only five matched seeds, the empirical conclusions below are based on effect patterns and consistency rather than significance declarations.

\subsection{Synthetic label corruption}
\begin{table*}[t]
\caption{Digits test performance with 40\% symmetric corruption applied only to training labels. Lower NLL, Brier score, and ECE are better. Values are mean $\pm$ sample standard deviation over five seeds.}
\label{tab:digits-noise}
\centering\small
\begin{tabular}{lcccc}
\toprule
Loss & Accuracy & NLL & Brier score & ECE\\
\midrule
SCE & 0.936 $\pm$ 0.013 & 0.246 $\pm$ 0.069 & 0.101 $\pm$ 0.020 & 0.037 $\pm$ 0.003\\
GCE & 0.934 $\pm$ 0.016 & 0.244 $\pm$ 0.063 & 0.101 $\pm$ 0.018 & 0.027 $\pm$ 0.008\\
APL & 0.928 $\pm$ 0.019 & 0.261 $\pm$ 0.082 & 0.109 $\pm$ 0.023 & 0.028 $\pm$ 0.005\\
MAE & 0.926 $\pm$ 0.021 & 0.266 $\pm$ 0.085 & 0.112 $\pm$ 0.025 & 0.028 $\pm$ 0.003\\
BTL & 0.923 $\pm$ 0.014 & 0.470 $\pm$ 0.039 & 0.183 $\pm$ 0.014 & 0.211 $\pm$ 0.019\\
CAPM & 0.921 $\pm$ 0.017 & 0.684 $\pm$ 0.035 & 0.279 $\pm$ 0.016 & 0.350 $\pm$ 0.028\\
APMS & 0.921 $\pm$ 0.017 & 0.705 $\pm$ 0.032 & 0.290 $\pm$ 0.015 & 0.368 $\pm$ 0.031\\
HPG & 0.920 $\pm$ 0.018 & 0.695 $\pm$ 0.034 & 0.285 $\pm$ 0.016 & 0.360 $\pm$ 0.030\\
CE & 0.915 $\pm$ 0.021 & 0.762 $\pm$ 0.035 & 0.316 $\pm$ 0.013 & 0.391 $\pm$ 0.025\\
\bottomrule
\end{tabular}
\end{table*}

Under 40\% symmetric corruption on Digits, SCE, GCE, APL, and MAE have higher mean accuracy and substantially lower NLL and Brier score than cross-entropy and the three structured objectives (Table~\ref{tab:digits-noise}). This result is consistent with those methods being proposed or commonly used for noisy-label robustness; it does not follow from boundedness or strict propriety alone. CAPM, HPG, and APMS modestly exceed cross-entropy in mean accuracy but remain behind the selected robust-loss baselines in this setting.

\begin{figure}[t]
\centering
\includegraphics[width=.86\linewidth]{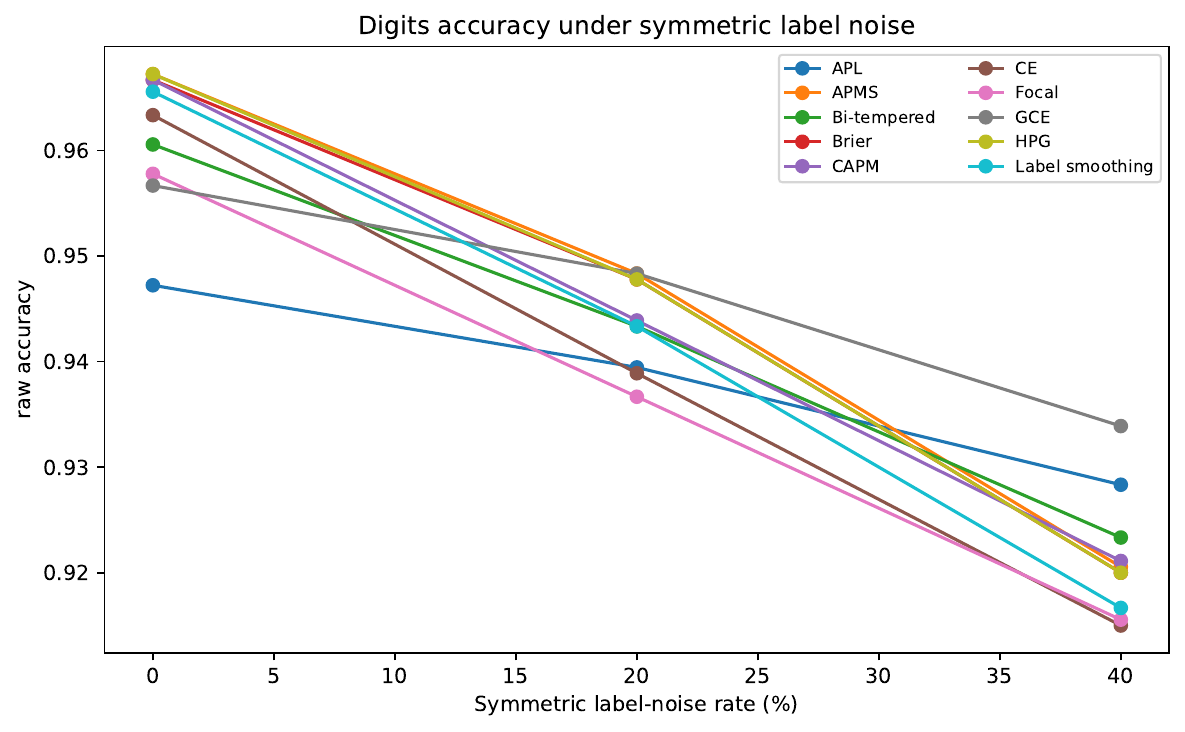}
\caption{Digits test accuracy under symmetric training-label corruption. The horizontal axis is the training-label corruption rate, validation and test labels are clean, and the vertical axis is ordinary test accuracy. Points summarize five retained seeds for each loss, with uncertainty drawn from those retained runs; higher values are better.}
\label{fig:noise-accuracy}
\end{figure}

\begin{figure}[t]
\centering
\includegraphics[width=.86\linewidth]{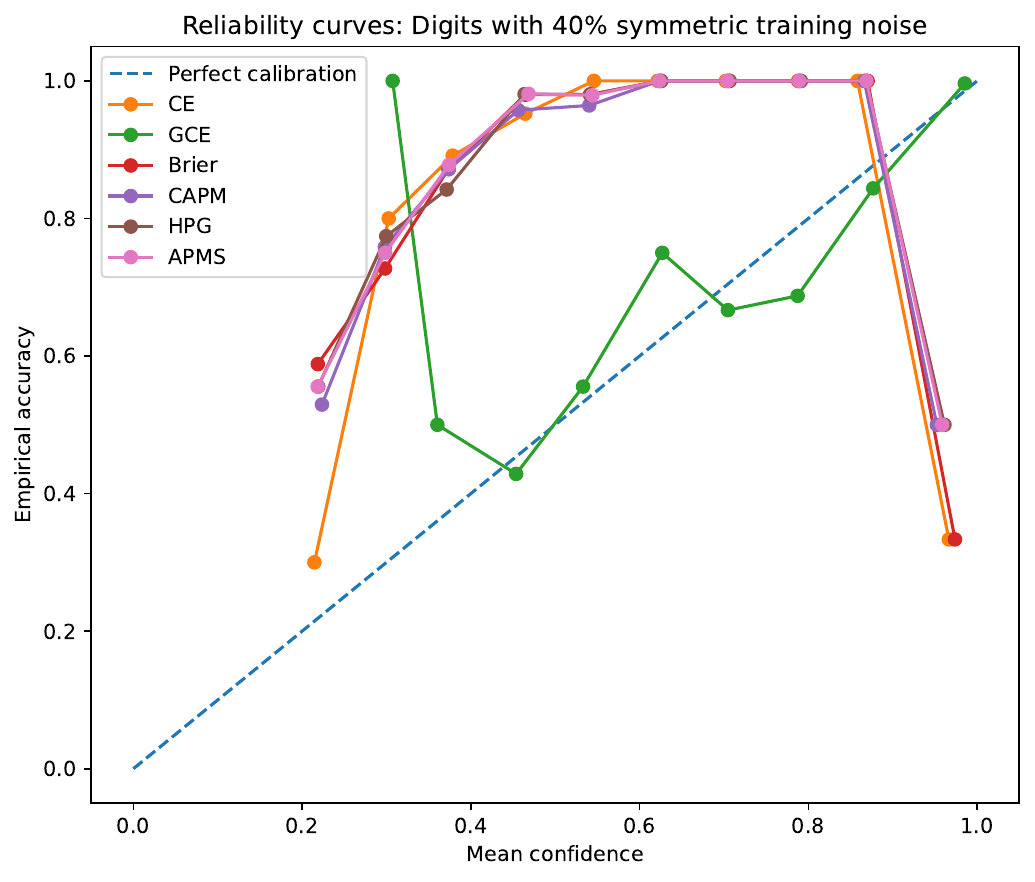}
\caption{Reliability curves on Digits after training with 40\% symmetric label corruption. The horizontal axis is mean top-class confidence and the vertical axis is empirical accuracy within confidence bins; the diagonal is perfect calibration. Curves are descriptive finite-test-set summaries and are not population calibration guarantees.}
\label{fig:reliability}
\end{figure}

The reliability diagram in Fig.~\ref{fig:reliability} illustrates that raw confidence behavior differs substantially across objectives under the same corruption level. The figure is descriptive: calibration curves from a small test set and five fitted models should not be interpreted as population calibration proofs.

\subsection{Long-tailed training}
\begin{table*}[t]
\caption{Performance on the balanced test split after training on a synthetic 30:1 long-tailed sample. Low ECE is not sufficient evidence of useful probabilities when balanced accuracy is low; NLL and Brier score are therefore reported jointly.}
\label{tab:longtail}
\centering\small
\resizebox{\linewidth}{!}{%
\begin{tabular}{lcccc}
\toprule
Loss & Balanced accuracy & NLL & Brier score & ECE\\
\midrule
Logit adjustment & 0.629 $\pm$ 0.012 & 1.128 $\pm$ 0.035 & 0.517 $\pm$ 0.012 & 0.051 $\pm$ 0.012\\
Balanced Softmax & 0.629 $\pm$ 0.012 & 1.128 $\pm$ 0.035 & 0.517 $\pm$ 0.012 & 0.051 $\pm$ 0.011\\
CB-CE & 0.628 $\pm$ 0.019 & 1.158 $\pm$ 0.043 & 0.514 $\pm$ 0.015 & 0.058 $\pm$ 0.006\\
LDAM & 0.613 $\pm$ 0.008 & 1.663 $\pm$ 0.024 & 0.719 $\pm$ 0.010 & 0.376 $\pm$ 0.010\\
CE & 0.164 $\pm$ 0.029 & 2.295 $\pm$ 0.007 & 0.898 $\pm$ 0.001 & 0.042 $\pm$ 0.012\\
CAPM & 0.163 $\pm$ 0.026 & 2.295 $\pm$ 0.008 & 0.898 $\pm$ 0.002 & 0.035 $\pm$ 0.020\\
APMS & 0.162 $\pm$ 0.028 & 2.295 $\pm$ 0.007 & 0.898 $\pm$ 0.002 & 0.037 $\pm$ 0.016\\
HPG & 0.162 $\pm$ 0.028 & 2.295 $\pm$ 0.008 & 0.898 $\pm$ 0.002 & 0.036 $\pm$ 0.016\\
\bottomrule
\end{tabular}%
}
\end{table*}

Balanced Softmax, logit adjustment, class-balanced cross-entropy, and LDAM have higher reported balanced accuracy than the generic objectives on the balanced test split (Table~\ref{tab:longtail}). Balanced Softmax and training-time logit adjustment have matching rounded values in this implementation at $\tau=1$ because their additive log-count and log-prior terms differ only by a class-independent constant. The structured curvature in CAPM does not compensate for the training-prior shift in this experiment: CAPM, HPG, APMS, and cross-entropy remain near 0.16 balanced accuracy. For $K=10$, a uniform forecast has NLL $\log 10$ and multiclass Brier score $0.9$. The observed values are close to both, so the small ECE values are consistent with near-uniform, low-information predictions rather than useful discrimination.

\begin{figure}[t]
\centering
\includegraphics[width=.88\linewidth]{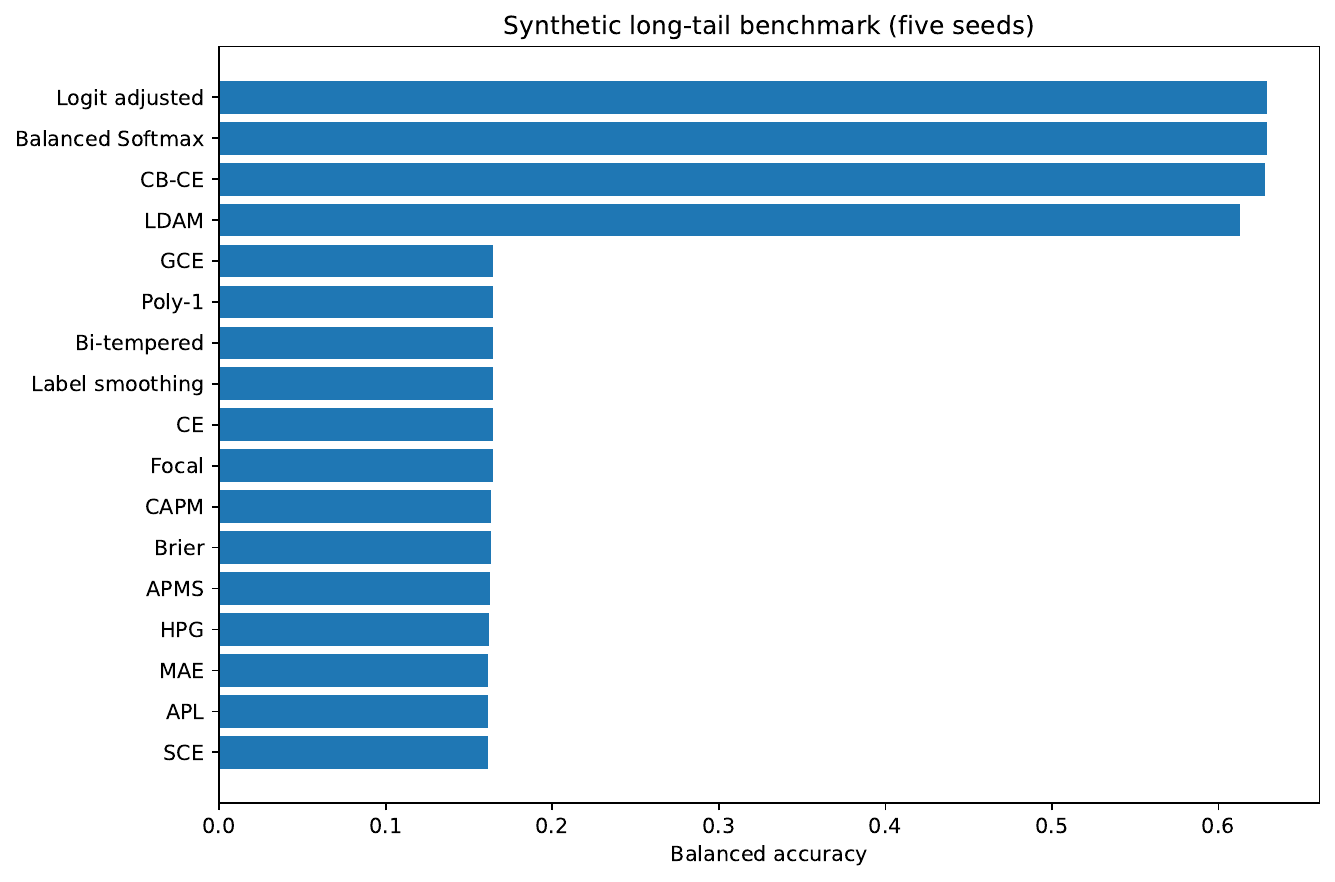}
\caption{Balanced accuracy on the balanced synthetic test split after 30:1 long-tailed training. Bars summarize five retained seeds; higher values are better. The plot separates prior-adjustment and margin/weighting methods from the generic losses in this particular synthetic-prior-shift design.}
\label{fig:longtail}
\end{figure}

\subsection{Calibration and perturbation sensitivity}
\begin{table}[t]
\caption{Digits clean-set NLL and ECE before and after validation-set temperature scaling. Means over five seeds.}
\label{tab:calibration}
\centering\small
\begin{tabular}{lrrrr}
\toprule
Loss & Raw NLL & Cal. NLL & Raw ECE & Cal. ECE\\
\midrule
CE & 0.136 & 0.124 & 0.038 & 0.021\\
Brier & 0.148 & 0.125 & 0.055 & 0.023\\
CAPM & 0.145 & 0.126 & 0.050 & 0.024\\
HPG & 0.147 & 0.125 & 0.053 & 0.022\\
APMS & 0.150 & 0.126 & 0.054 & 0.023\\
\bottomrule
\end{tabular}
\end{table}

Validation-set temperature scaling reduces NLL and ECE for the selected methods on clean Digits (Table~\ref{tab:calibration}) without changing predicted labels. This post-hoc improvement does not establish that one training loss is intrinsically better calibrated, because the temperature is estimated after training and uses finite validation data.

\begin{figure}[t]
\centering
\includegraphics[width=.88\linewidth]{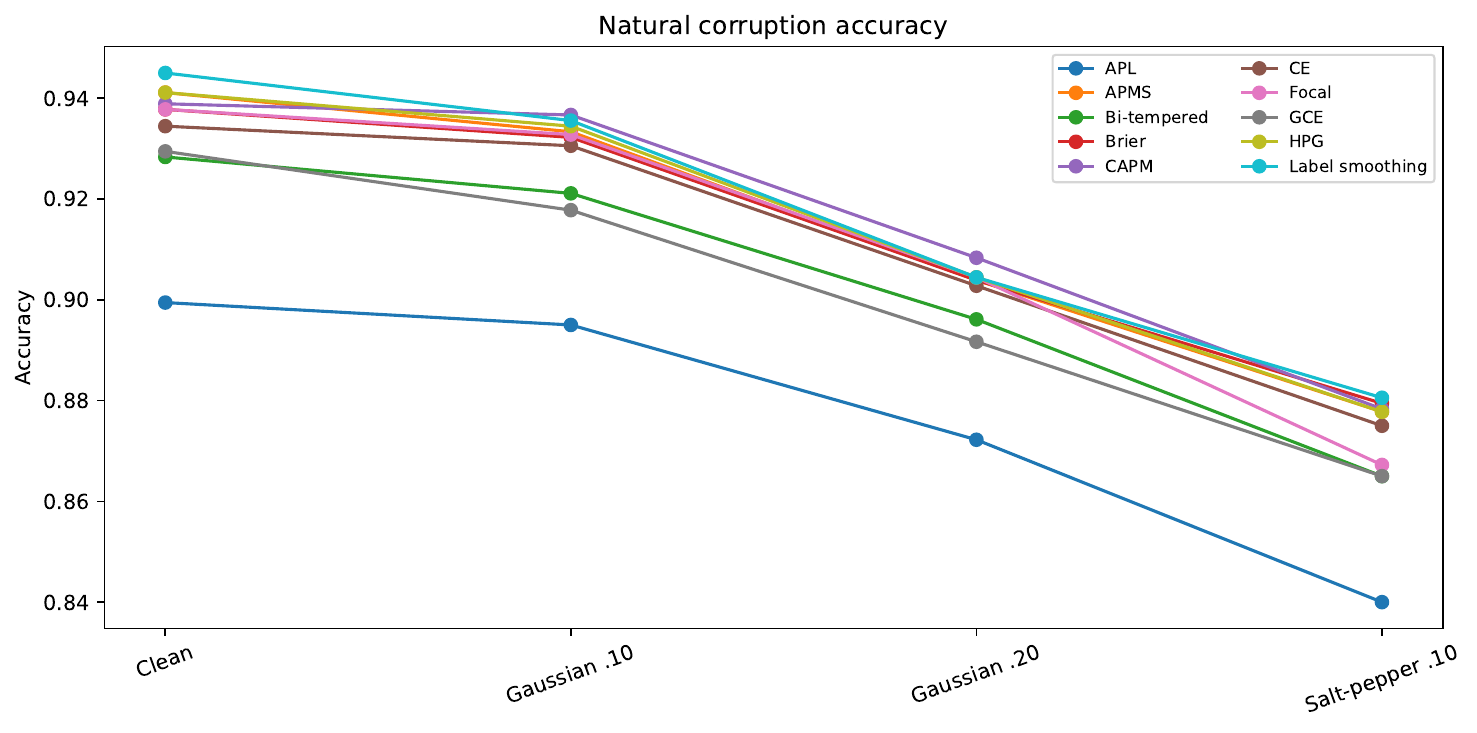}
\caption{Accuracy of clean-trained Digits models under synthetic input corruptions. The conditions are the uncorrupted test set, Gaussian noise with standard deviations 0.10 and 0.20 on $[0,1]$-scaled pixels, and salt-and-pepper corruption with fraction 0.10. The vertical axis is ordinary test accuracy; higher values are better.}
\label{fig:natural-corruption}
\end{figure}

\begin{figure*}[t]
\centering
\begin{minipage}{.49\textwidth}
\centering\includegraphics[width=\linewidth]{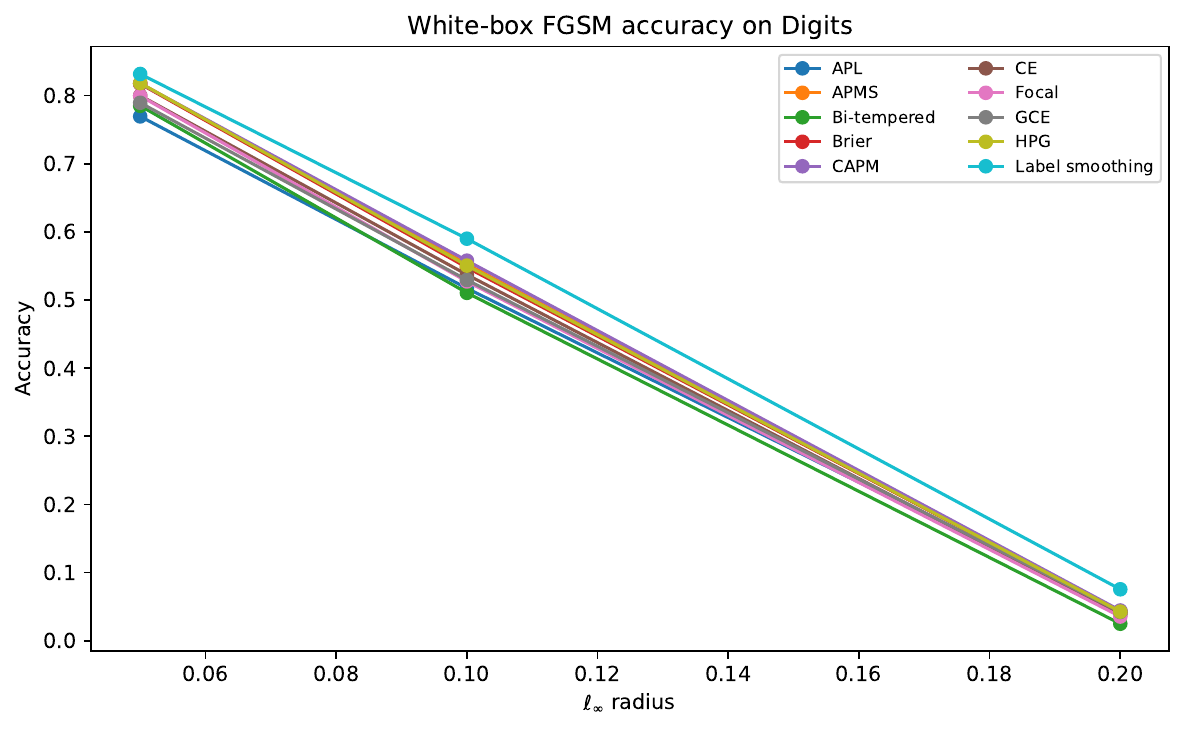}
\end{minipage}\hfill
\begin{minipage}{.49\textwidth}
\centering\includegraphics[width=\linewidth]{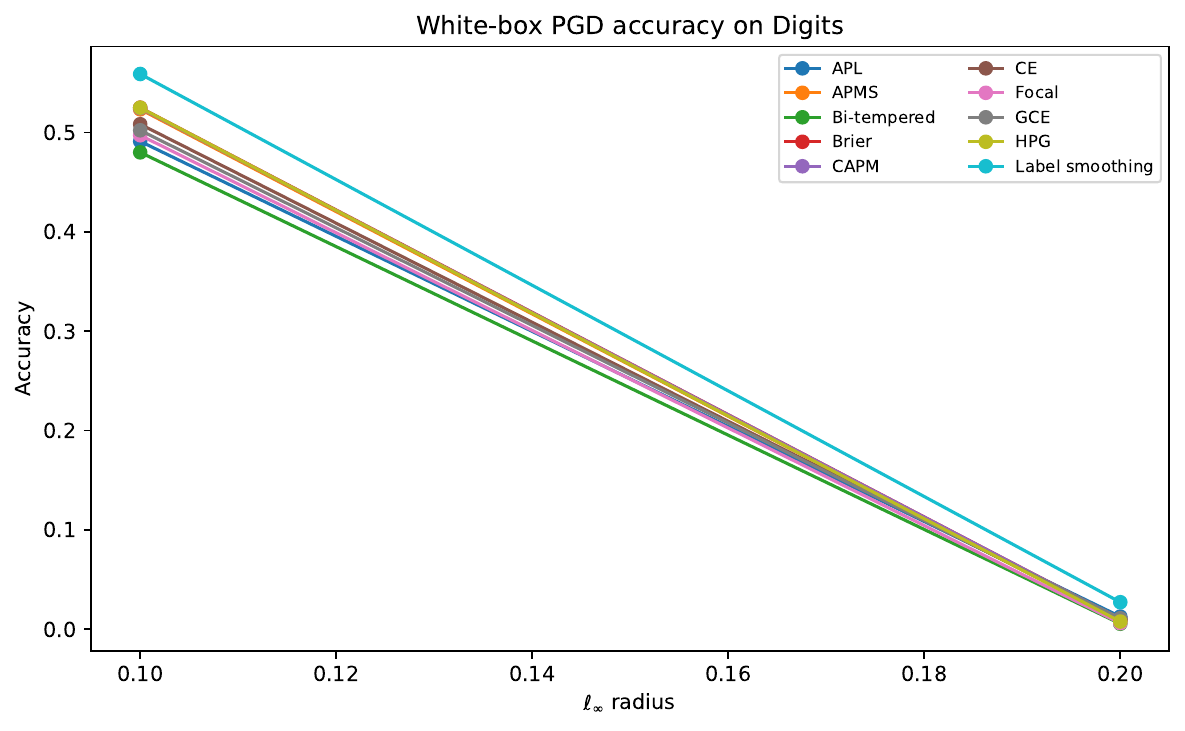}
\end{minipage}
\caption{Accuracy of clean-trained Digits models under white-box cross-entropy-driven FGSM and projected-gradient attacks. The left panel reports FGSM accuracy and the right panel reports projected-gradient accuracy as the perturbation radius increases. These finite attack evaluations are local sensitivity probes and are not certified robustness results.}
\label{fig:attacks}
\end{figure*}

The input-perturbation probes show small descriptive differences among clean-trained Digits models under Gaussian noise, salt-and-pepper corruption, FGSM, and projected-gradient attacks (Figs.~\ref{fig:natural-corruption} and~\ref{fig:attacks}). Since the models were not adversarially trained and the attack suite is limited, these observations support only a local sensitivity comparison.

\subsection{Ablation results}
The candidate-specific ablations do not show a large, consistent benefit from the added structures. On the synthetic-confusion task, CAPM variants without graph or tail terms can match or exceed the full configuration. Increasing the number of HPG ridge terms does not produce monotonic gains across the evaluated cells. Varying the initial APMS coefficient changes mean accuracy only slightly and inconsistently. These negative results are important because they weaken the interpretation that the graph, ridge, or temporary margin component is responsible for the candidate performance.

\begin{figure*}[t]
\centering
\begin{minipage}{.32\textwidth}\centering
\includegraphics[width=\linewidth]{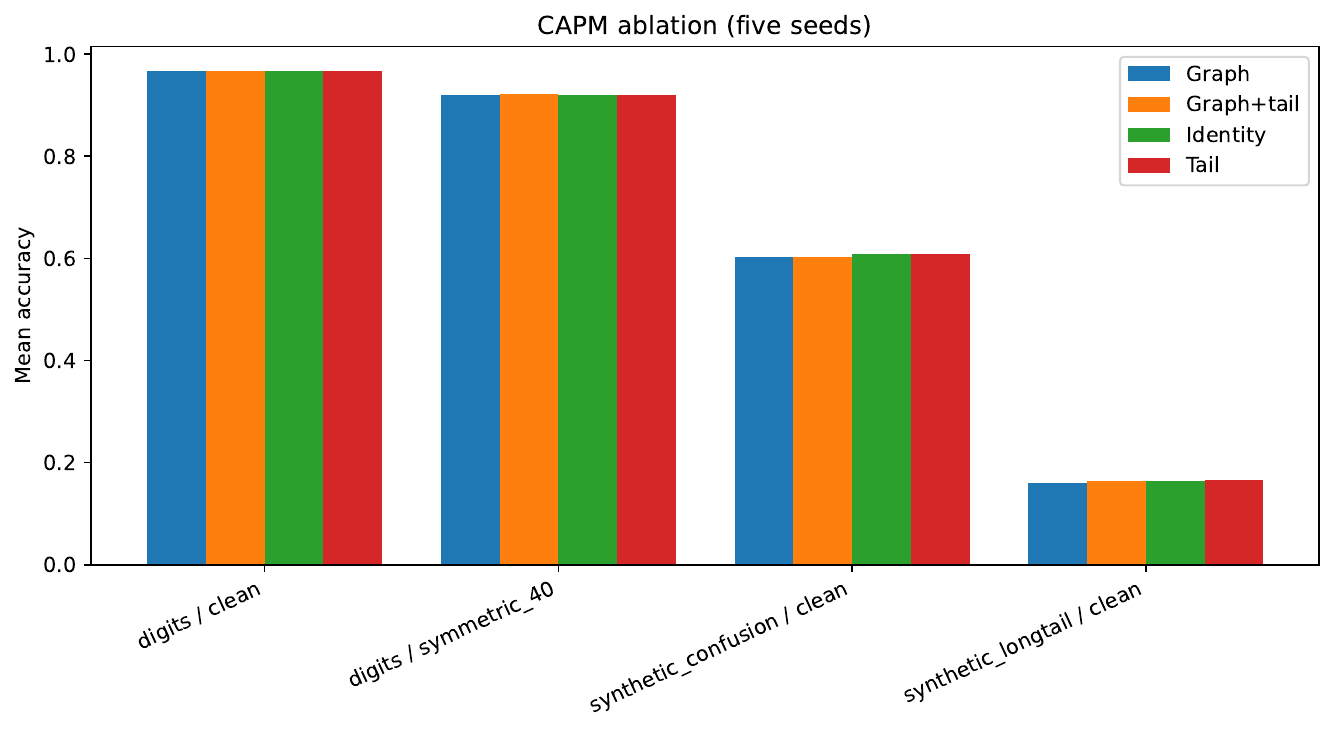}
\end{minipage}\hfill
\begin{minipage}{.32\textwidth}\centering
\includegraphics[width=\linewidth]{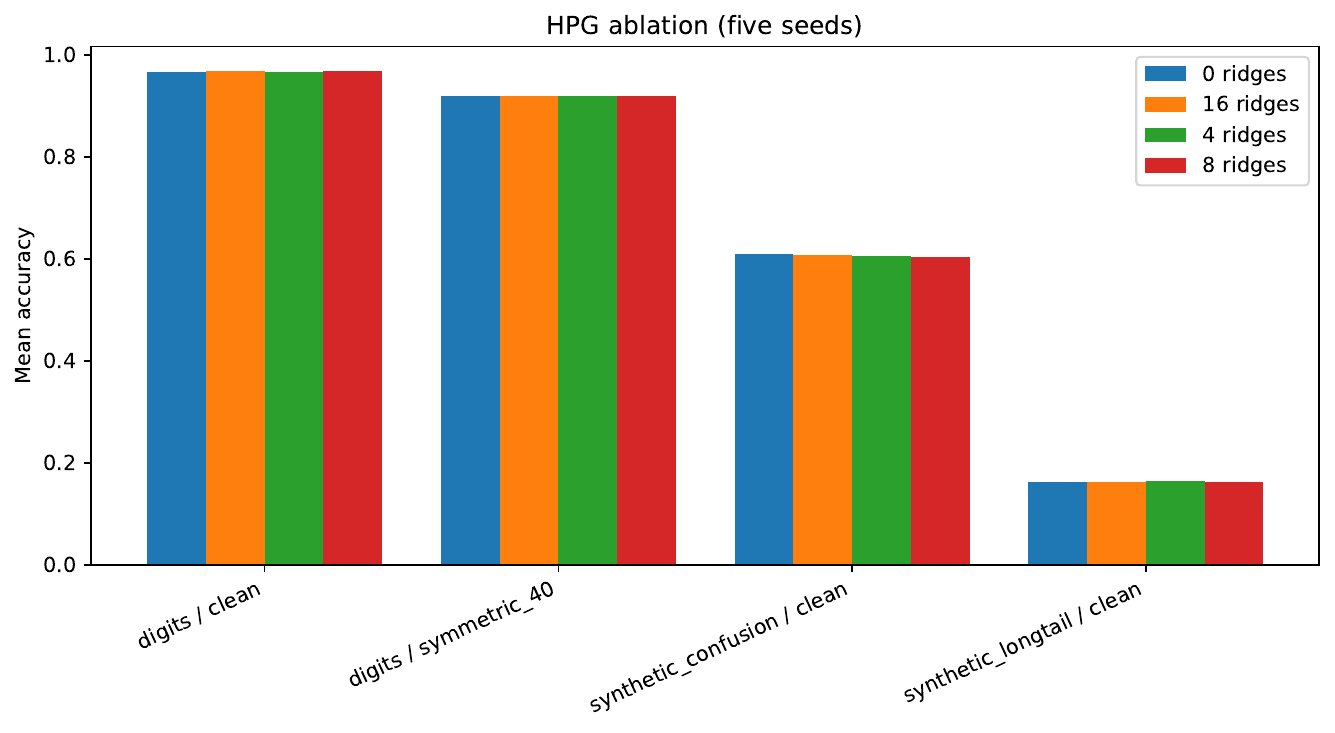}
\end{minipage}\hfill
\begin{minipage}{.32\textwidth}\centering
\includegraphics[width=\linewidth]{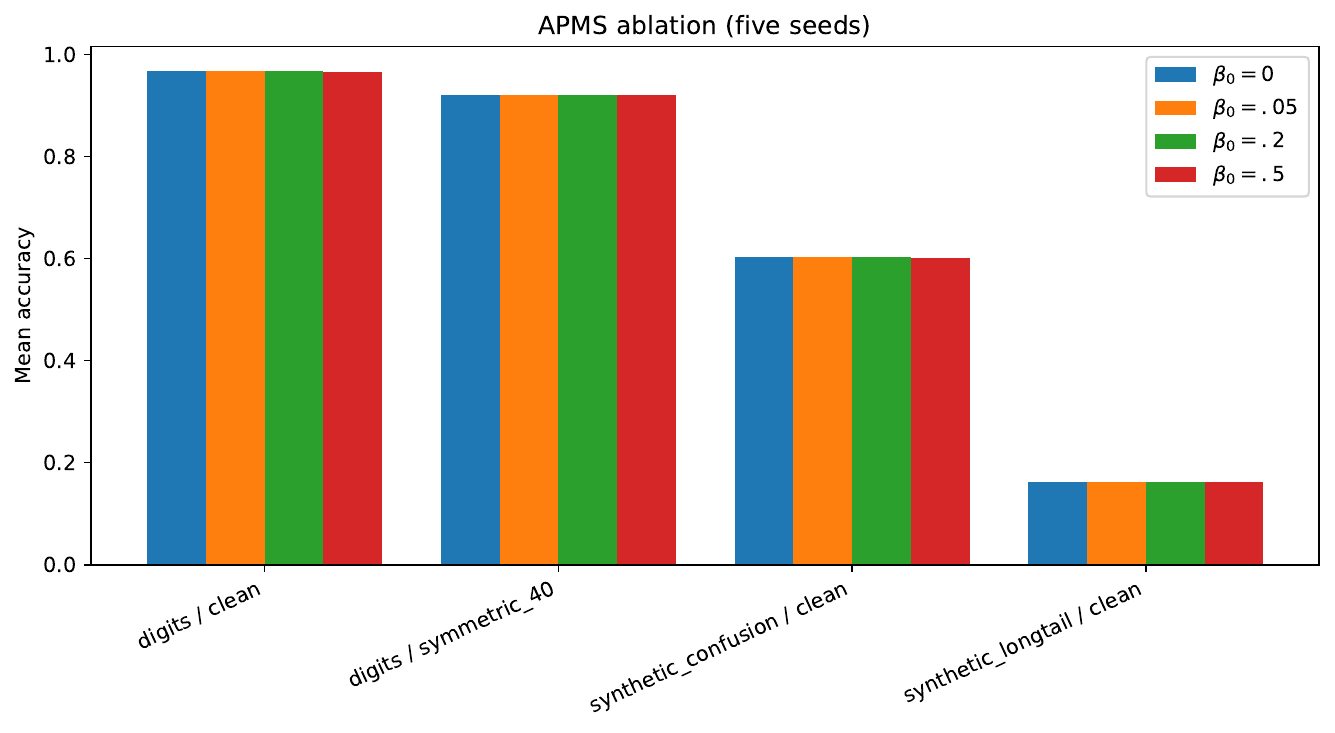}
\end{minipage}
\caption{Candidate-specific accuracy ablations. The left panel compares CAPM geometry variants, the middle panel compares HPG ridge counts, and the right panel compares APMS initial margin coefficients across the retained ablation cells. Values are descriptive five-seed means; higher accuracy is better, and no statistical significance claim is attached to these differences.}
\label{fig:ablations}
\end{figure*}

\section{Discussion}\label{sec:discussion}

The theoretical and empirical results distinguish population identification from finite-sample robustness. CAPM and HPG are strictly proper because their generators are strongly convex, and their excess conditional risks are controlled by squared probability error. These properties establish a well-defined population target and bounded probability-space gradients. They do not imply resistance to corrupted labels or class-prior shift. The noisy-label experiments support this distinction in the evaluated setting: the selected noisy-label baselines outperform the proper structured candidates under 40\% symmetric corruption on Digits.

The long-tail result is similarly instructive. Adding a class-frequency-dependent diagonal to a proper quadratic geometry changes curvature but does not implement the prior correction used by Balanced Softmax or logit adjustment. In the present synthetic setting, that distinction is large enough to separate low balanced accuracy near 0.16 from balanced accuracy above 0.62. A future class-aware proper score intended for long-tail learning would need a statistical treatment of prior shift rather than curvature modification alone.

HPG provides a smooth, globally controlled alternative to a purely quadratic generator, but the ablations do not establish that its log-cosh ridges improve predictive performance. APMS has a clear asymptotic relationship to its proper core, yet finite optimization can select a checkpoint while the margin coefficient remains positive. The displacement bounds describe conditional minimizers as $\beta$ decreases; they do not prove that stochastic neural optimization follows that path or improves generalization.

Overall, the proofs support the mathematical validity of the formulations, while the experiments do not support a claim that they replace established losses. Their most defensible role in this paper is as controlled examples for studying how proper-score curvature and temporary margin shaping affect optimization and probability quality.

\section{Limitations}\label{sec:limitations}

The empirical study uses small tabular and low-resolution image-feature datasets with a compact multilayer perceptron. Conclusions may not transfer to convolutional networks, transformers, or large natural datasets. The label corruption is synthetic and excludes instance-dependent, annotator-dependent, and open-set noise. The long-tail experiment uses a generated distribution and a balanced test set; natural long-tailed recognition can involve representation and domain effects not present here.

Five seeds provide limited precision and make exact paired rank tests underpowered. A common learning rate controls compute but cannot guarantee equal optimization quality for every loss. Although a three-rate sensitivity analysis is retained in the supplementary results, exhaustive per-loss tuning was not performed. ECE is bin-dependent, and temperature scaling reuses the validation split used for checkpoint selection rather than a fully nested calibration partition.

The adversarial tests use a small attack set, no adversarial training, and no certified verification. The logit-gradient bound in Eq.~\eqref{eq:logit-gradient} does not control the network input Jacobian. Finally, the paper does not establish historical priority for the exact HPG or APMS formulas; it evaluates and analyzes the stated constructions without an absolute novelty claim.

\section{Conclusion}\label{sec:conclusion}

This study examined two structured proper scoring rules and an annealed margin-augmented objective for multiclass neural classification. CAPM instantiates a class-structured quadratic Bregman geometry. HPG adds bounded-curvature log-cosh ridges while retaining strict propriety. APMS temporarily perturbs HPG and has a conditional minimizer whose distance from the true class distribution is bounded by the stated quantities that vanish with the margin coefficient.

The controlled experiments do not identify a universal winner. The candidates are close to cross-entropy in several clean and noisy cells, but the selected noisy-label baselines perform better on Digits with 40\% symmetric label noise and explicit prior-adjustment methods perform much better in the 30:1 synthetic long-tail experiment. Ablations provide no consistent evidence that the added graph, ridge, or margin structures are beneficial in the evaluated settings. The appropriate conclusion is therefore limited: the formulations satisfy the stated propriety and bound results and are empirically testable, but broader utility requires larger datasets, stronger architectures, realistic noise, and adequately powered comparisons.

\clearpage

\printbibliography

\clearpage

\appendix
\section{Proofs}\label{app:proofs}

\subsection{Proof of Lemma~\ref{lem:regret}}
Linearity of expectation gives
\begin{align}
R_F(\eta,p)
&=\sum_y\eta_yF(e_y)-F(p)-\langle\nabla F(p),\eta-p\rangle,\\
R_F(\eta,\eta)&=\sum_y\eta_yF(e_y)-F(\eta).
\end{align}
Subtracting yields Eq.~\eqref{eq:conditional-regret}. Strict convexity makes $D_F(\eta,p)=0$ if and only if $p=\eta$. For twice differentiable $F$, writing $d=q-p$ gives
\begin{equation}
D_F(q,p)=\int_0^1(1-t)d^\top H_F(p+td)d\,dt.
\end{equation}
The spectral bounds $mI\preceq H_F\preceq MI$ and $\int_0^1(1-t)dt=1/2$ yield Eq.~\eqref{eq:bregman-quadratic}.

\subsection{Softmax Jacobian bound}
For a unit vector $v$,
\begin{equation}
v^\top\{\operatorname{Diag}(p)-pp^\top\}v
=\operatorname{Var}_{I\sim p}(v_I).
\end{equation}
Popoviciu's inequality bounds this variance by one quarter of the squared range of the coordinates of $v$. A unit vector has coordinate range at most $\sqrt2$, so the operator norm is at most $1/2$. The temperature-scaled softmax contributes the factor $1/T$, which yields Eq.~\eqref{eq:logit-gradient} after applying Eq.~\eqref{eq:prob-gradient}.

\subsection{Proof of Proposition~\ref{prop:capm}}
The Hessian of $F_{\mathrm C}$ is the constant positive-definite matrix $A$. Lemma~\ref{lem:regret} gives strict propriety and the stated conditional regret. Rayleigh-quotient bounds give Eq.~\eqref{eq:bregman-quadratic}. Since $\|e_y-p\|_2^2\le2$ on the simplex,
\begin{equation}
\ell_{\mathrm C}(p,y)\le\frac12M\|e_y-p\|_2^2\le M.
\end{equation}
Equation~\eqref{eq:logit-gradient} follows from Eq.~\eqref{eq:prob-gradient} and the softmax Jacobian bound.

\subsection{Proof of Proposition~\ref{prop:hpg}}
Each ridge contributes
\begin{equation}
a_r\operatorname{sech}^2(v_r)w_rw_r^\top\succeq0
\end{equation}
to the unscaled Hessian and is bounded above by $a_r\|w_r\|_2^2I$. Adding the quadratic term and multiplying by $s$ proves $mI\preceq H_{F_{\mathrm H}}\preceq MI$. Lemma~\ref{lem:regret} gives strict propriety and the quadratic regret bounds, while $\|e_y-p\|_2^2\le2$ gives the loss-range bound.

For one ridge, differentiation of the Hessian in direction $h$ yields
\begin{equation}
-\frac{2sa_r}{\rho_r}\operatorname{sech}^2(v_r)\tanh(v_r)
(w_r^\top h)w_rw_r^\top.
\end{equation}
The maximum of $2t(1-t^2)$ on $t\in[0,1]$ is $4/(3\sqrt3)$. Taking operator norms and summing over ridges proves Eq.~\eqref{eq:hessian-lipschitz}.

\subsection{Proof of Theorem~\ref{thm:apms}}
Fix $a=p_y$. For this value of $a$, minimizing $m_\tau(p,y)$ is equivalent to maximizing the log-sum-exp term over the competing probabilities with total mass $1-a$. A convex function attains a maximum over a simplex at a vertex, so the smallest margin at fixed $a$ is
\begin{equation}
a-\tau\log\{\exp((1-a)/\tau)+K-2\}.
\end{equation}
Its derivative with respect to $a$ is
\begin{equation}
1+\frac{\exp((1-a)/\tau)}{\exp((1-a)/\tau)+K-2}>0,
\end{equation}
so the global minimum occurs at $a=0$. One competitor then has probability one and the remaining $K-2$ competitors have probability zero. This gives $m_{\min}$ in Eq.~\eqref{eq:margin-minimum}.

For the maximum margin at fixed $a$, the log-sum-exp term is minimized when the $K-1$ competing probabilities are equal. The largest margin at fixed $a$ is therefore
\begin{equation}
a-\tau\log\left\{(K-1)\exp\left(\frac{1-a}{(K-1)\tau}\right)\right\}
=\frac{Ka-1}{K-1}-\tau\log(K-1),
\end{equation}
which is increasing in $a$ and is maximized at $a=1$. This gives $m_{\max}$ in Eq.~\eqref{eq:margin-minimum}. Because softplus is increasing in $\kappa-m_\tau$, Eq.~\eqref{eq:penalty-bound} follows.

Optimality of $p_\beta^\star$ relative to $p=\eta$ gives
\begin{equation}
D_F(\eta,p_\beta^\star)\le \beta C_r.
\end{equation}
Strong convexity then yields Eq.~\eqref{eq:apms-sqrt-bound}. For the linear bound,
\begin{equation}
\nabla_pm_\tau=e_y-q_{-y},
\end{equation}
where $q_{-y}$ is a softmax distribution supported on the competing classes. Hence $\|\nabla_pm_\tau\|_2\le\sqrt2$. The derivative of the outer softplus with respect to $m_\tau$ has magnitude at most one, so the norm of the conditional penalty gradient is at most $\sqrt2$. The Bregman identity gives $\nabla_pR_0(\eta,p)=H_F(p)(p-\eta)$, and therefore
\begin{equation}
\langle\nabla_pR_0(\eta,p_\beta^\star),p_\beta^\star-\eta\rangle\ge m\|p_\beta^\star-\eta\|_2^2.
\end{equation}
Combining this inequality with the first-order variational inequality at $p_\beta^\star$ gives
\begin{equation}
m\|p_\beta^\star-\eta\|_2^2
\le\beta\sqrt2\|p_\beta^\star-\eta\|_2,
\end{equation}
which proves Eq.~\eqref{eq:apms-linear-bound}.

\subsection{Why APMS with a positive margin coefficient is not generally proper}
A binary counterexample is sufficient. Let $p=(x,1-x)$, $\eta=(0.8,0.2)$, $\kappa=0$, and $\nu=1$. In the binary case the two margins are $2x-1$ and $1-2x$. The derivative of the conditional penalty risk at the truthful report $x=0.8$ is
\begin{equation}
-2(0.8)\,\sigma(-0.6)+2(0.2)\,\sigma(0.6)\ne0,
\end{equation}
where $\sigma$ is the logistic function. The proper core has zero derivative at $x=0.8$, so adding any positive multiple of this penalty shifts the conditional stationary point. Thus positive-$\beta$ APMS is not a proper score in general.

\section{Supplementary empirical results}\label{app:supplementary}

\begin{figure}[t]
\centering
\includegraphics[width=.85\linewidth]{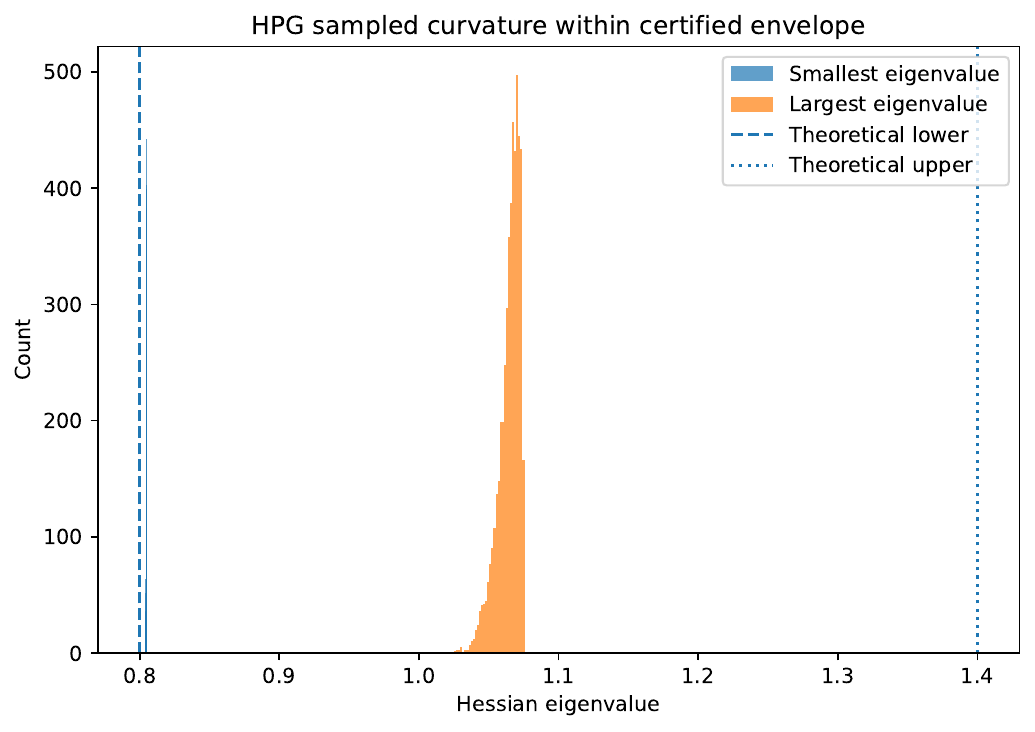}
\caption{HPG curvature diagnostic. The histogram shows sampled Hessian eigenvalues for the retained HPG construction, while the vertical reference lines mark the smallest and largest sampled eigenvalues and the analytical lower and upper bounds from Proposition~\ref{prop:hpg}. The plot is an implementation check of the sampled configuration; the guarantee is the analytical envelope, not the sample.}
\label{fig:hpg-curvature}
\end{figure}

\begin{figure}[t]
\centering
\includegraphics[width=.85\linewidth]{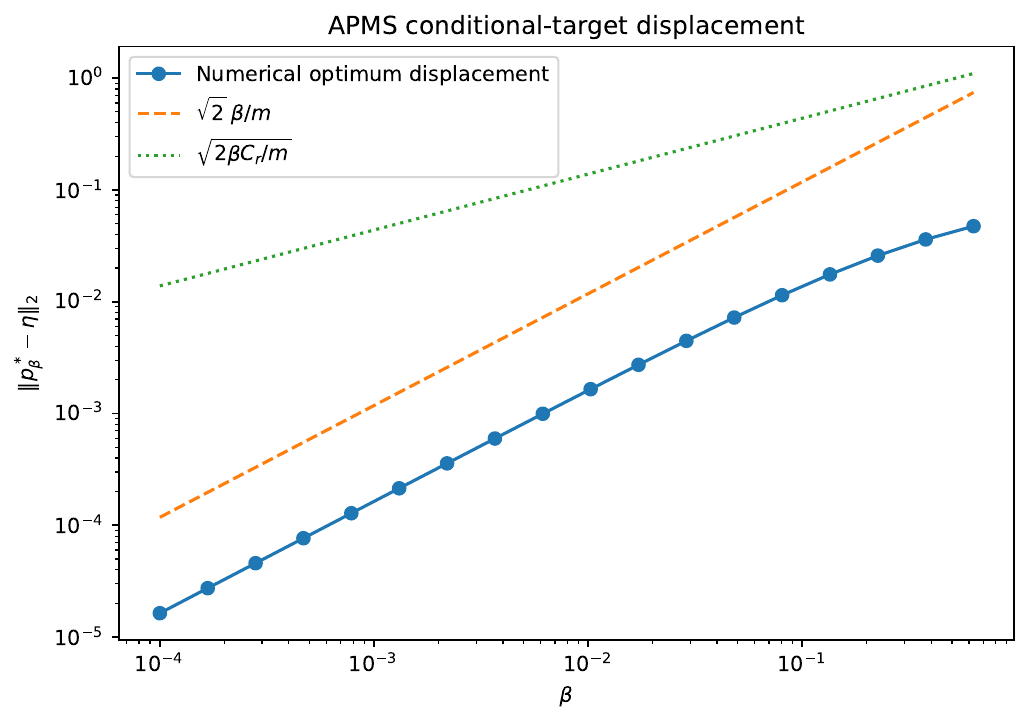}
\caption{APMS conditional-target displacement diagnostic. The numerical curve shows the optimized distance between the positive-$\beta$ APMS conditional minimizer and the true conditional distribution, and the two reference curves show the square-root and linear upper bounds in Theorem~\ref{thm:apms}. The display illustrates that the plotted displacement and both bounds decrease as the margin coefficient decreases.}
\label{fig:apms-displacement}
\end{figure}

\begin{figure}[t]
\centering
\includegraphics[width=.88\linewidth]{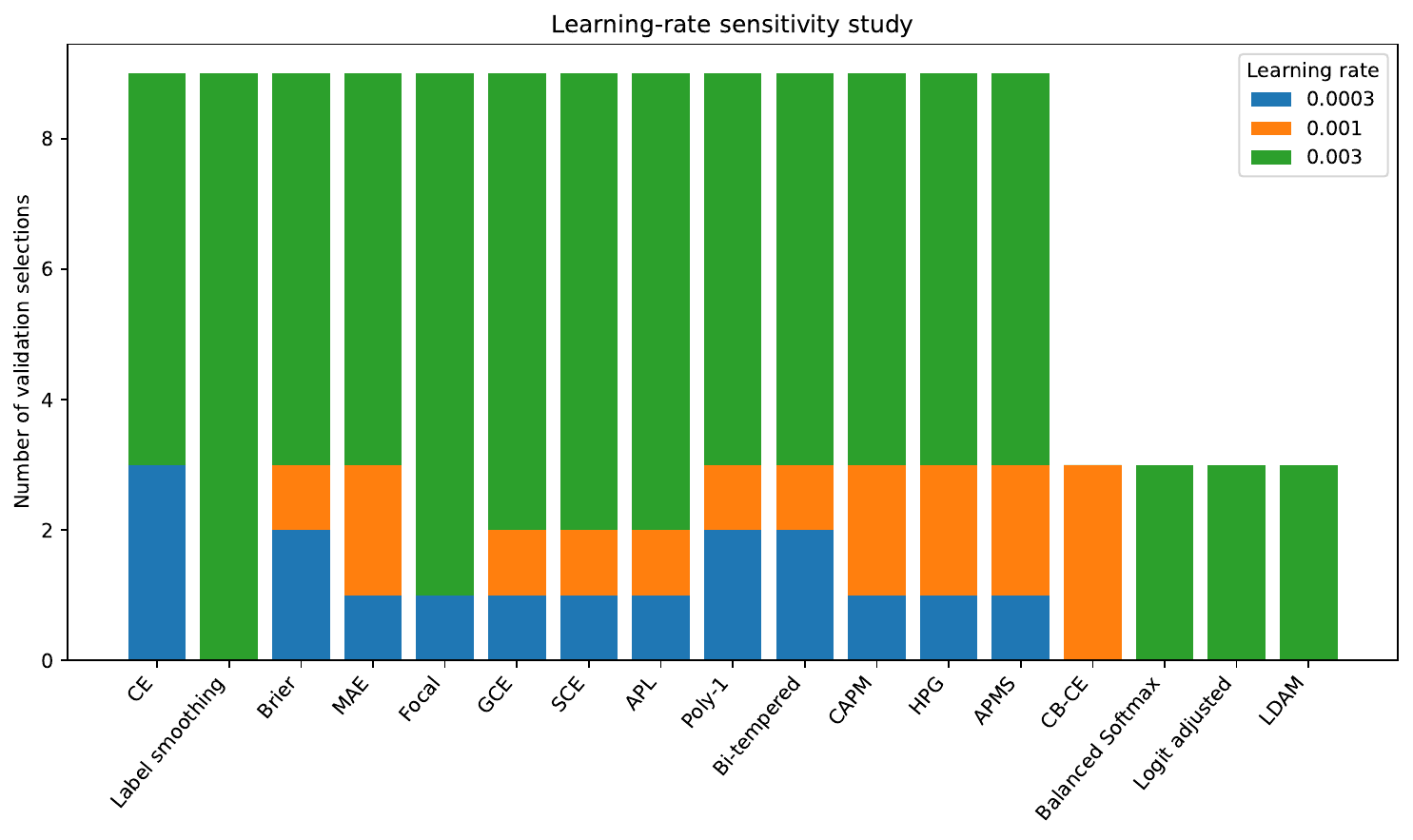}
\caption{Learning-rate sensitivity summary. For each loss, the bars count how often validation NLL selected one of the three candidate learning rates, $3\times10^{-4}$, $10^{-3}$, and $3\times10^{-3}$, across the retained sensitivity runs. The figure reports selection frequency only; it is not a test-set performance comparison.}
\label{fig:lr-sensitivity}
\end{figure}

The remaining supplementary figures collect the retained theoretical diagnostics and empirical summaries. Unless a caption states otherwise, empirical panels visualize retained run summaries and should be read as descriptive checks rather than additional inferential evidence. Accuracy and balanced accuracy are better when larger; NLL, ECE, mean rank, gradient norm, and runtime are better when smaller for the purposes stated in their captions.

\begin{figure*}[p]
\centering
\begin{minipage}{.32\textwidth}\centering
\includegraphics[width=\linewidth]{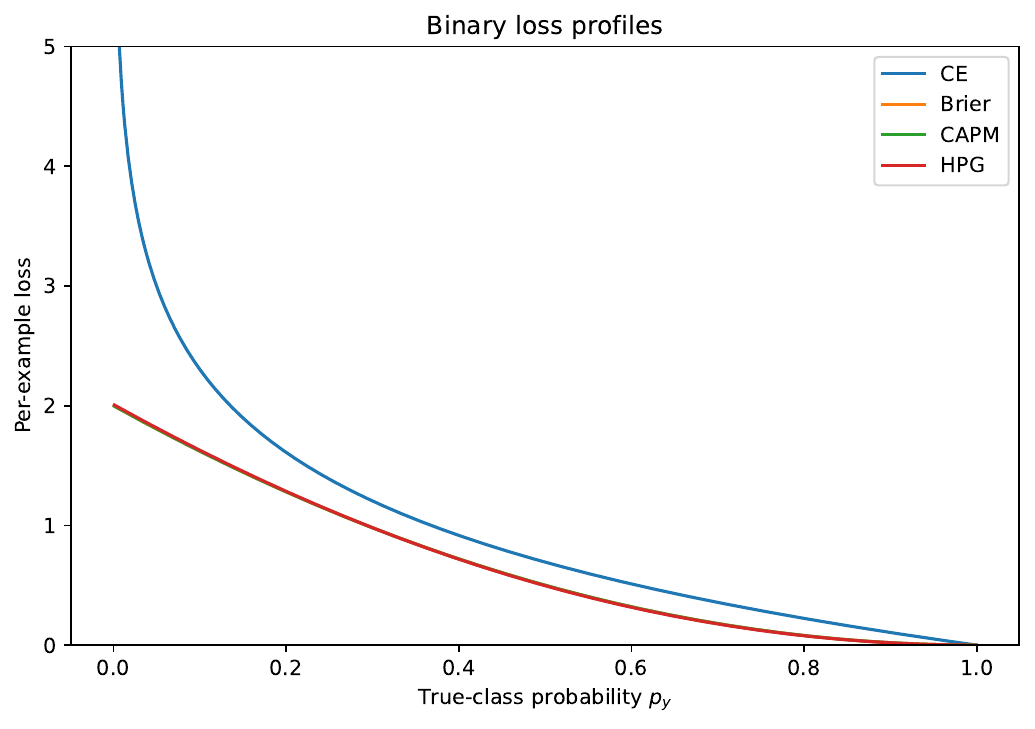}
\end{minipage}\hfill
\begin{minipage}{.32\textwidth}\centering
\includegraphics[width=\linewidth]{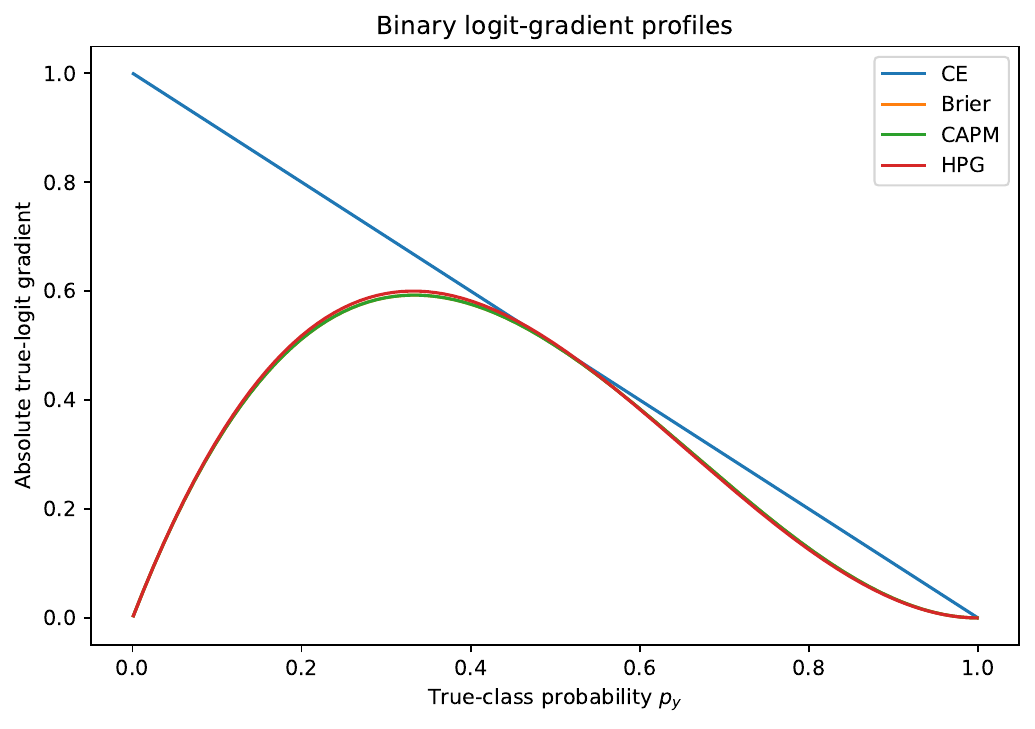}
\end{minipage}\hfill
\begin{minipage}{.32\textwidth}\centering
\includegraphics[width=\linewidth]{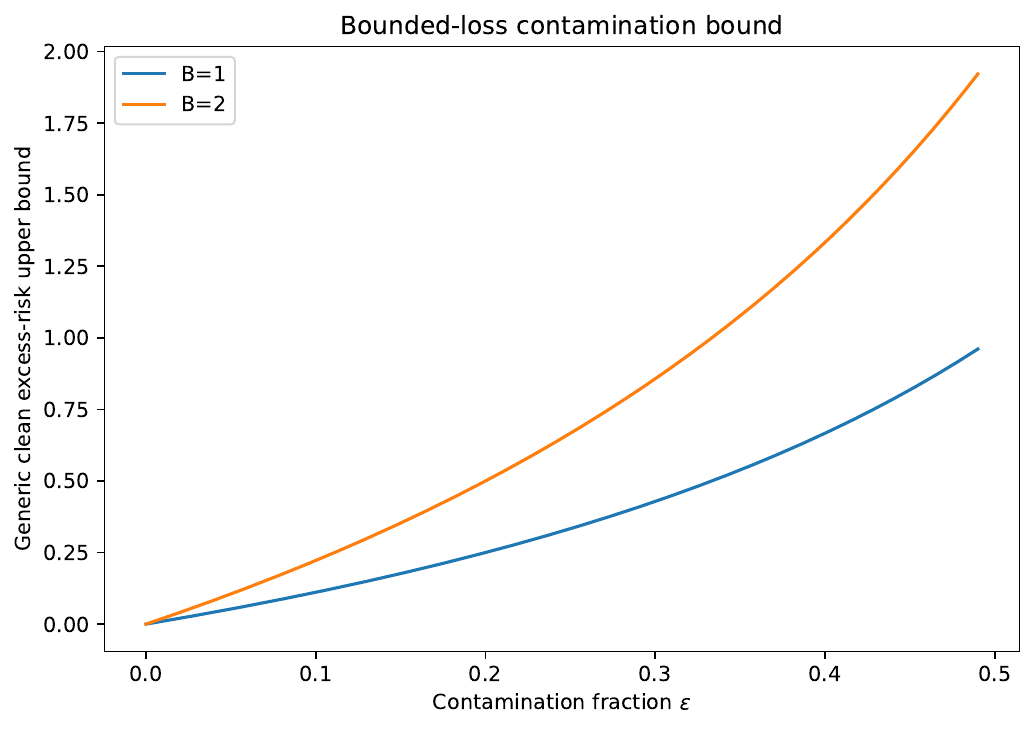}
\end{minipage}
\caption{Binary theoretical diagnostics. From left to right, the panels show per-example loss as the true-class probability varies, the absolute true-logit gradient as the true-class probability varies, and the generic bounded-loss contamination upper bound for two illustrative loss bounds. The first two panels compare CE, Brier, CAPM, and HPG in the binary setting; the third panel is a theoretical bound illustration rather than an empirical result.}
\label{fig:supp-theory-binary}
\end{figure*}

\begin{figure*}[p]
\centering
\begin{minipage}{.49\textwidth}\centering
\includegraphics[width=\linewidth]{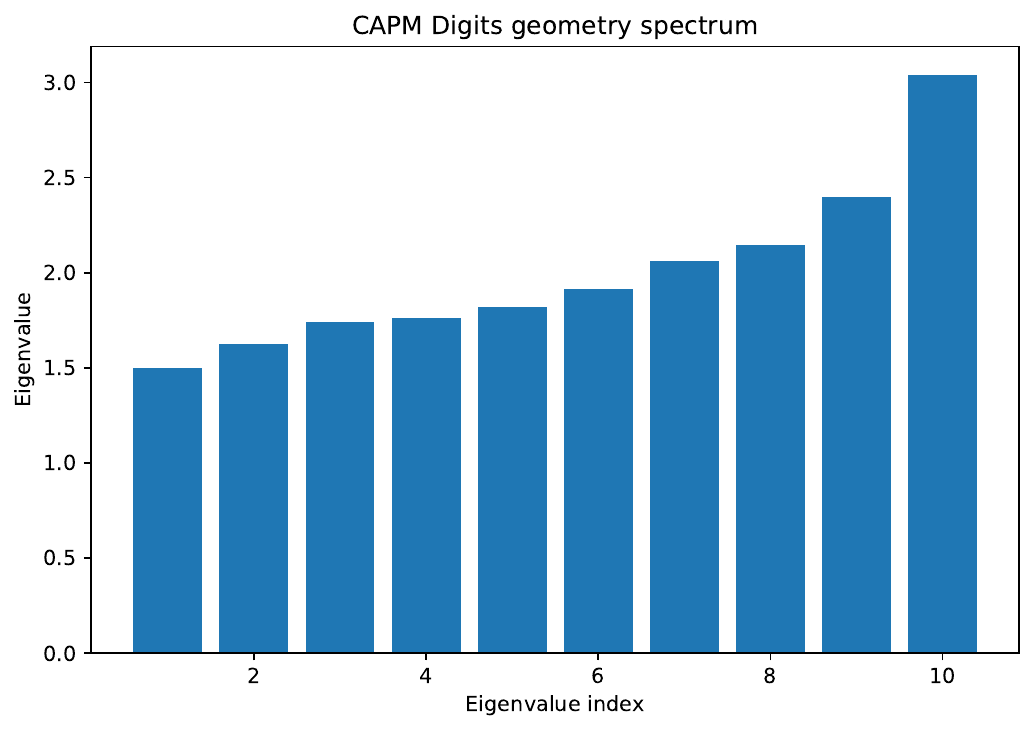}
\end{minipage}\hfill
\begin{minipage}{.49\textwidth}\centering
\includegraphics[width=\linewidth]{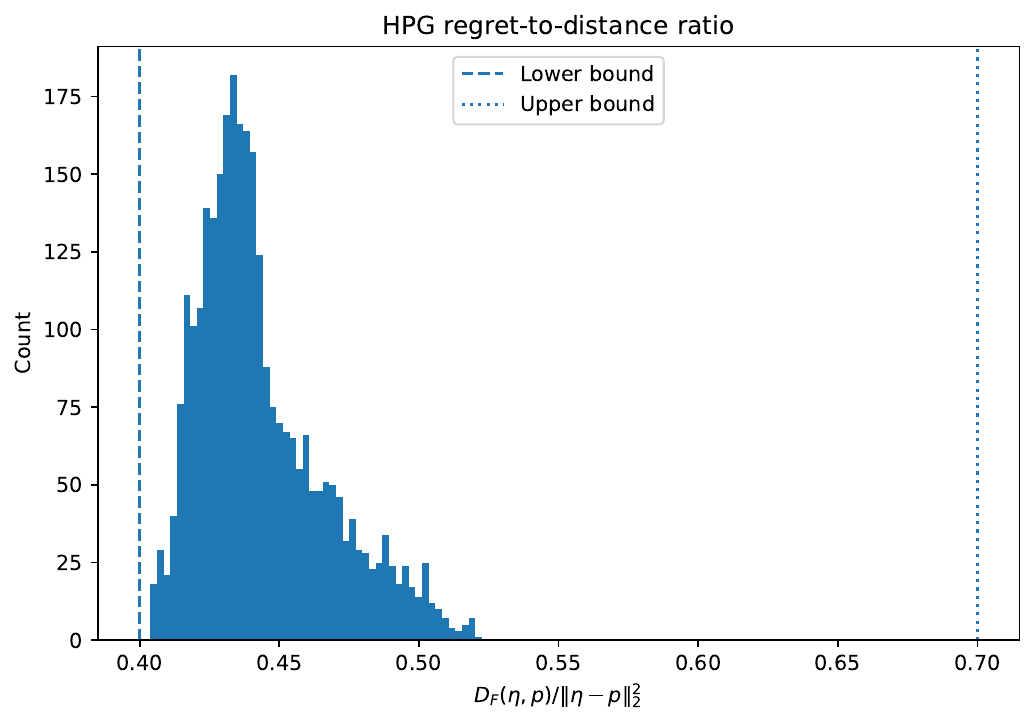}
\end{minipage}
\caption{Additional theory diagnostics. The left panel shows the eigenvalue spectrum of the CAPM geometry matrix used for Digits, confirming positive eigenvalues for the retained configuration. The right panel shows sampled values of the HPG regret-to-squared-distance ratio together with the lower and upper quadratic-regret bounds implied by the curvature envelope.}
\label{fig:supp-theory-spectrum-regret}
\end{figure*}

\begin{figure*}[p]
\centering
\begin{minipage}{.32\textwidth}\centering
\includegraphics[width=\linewidth]{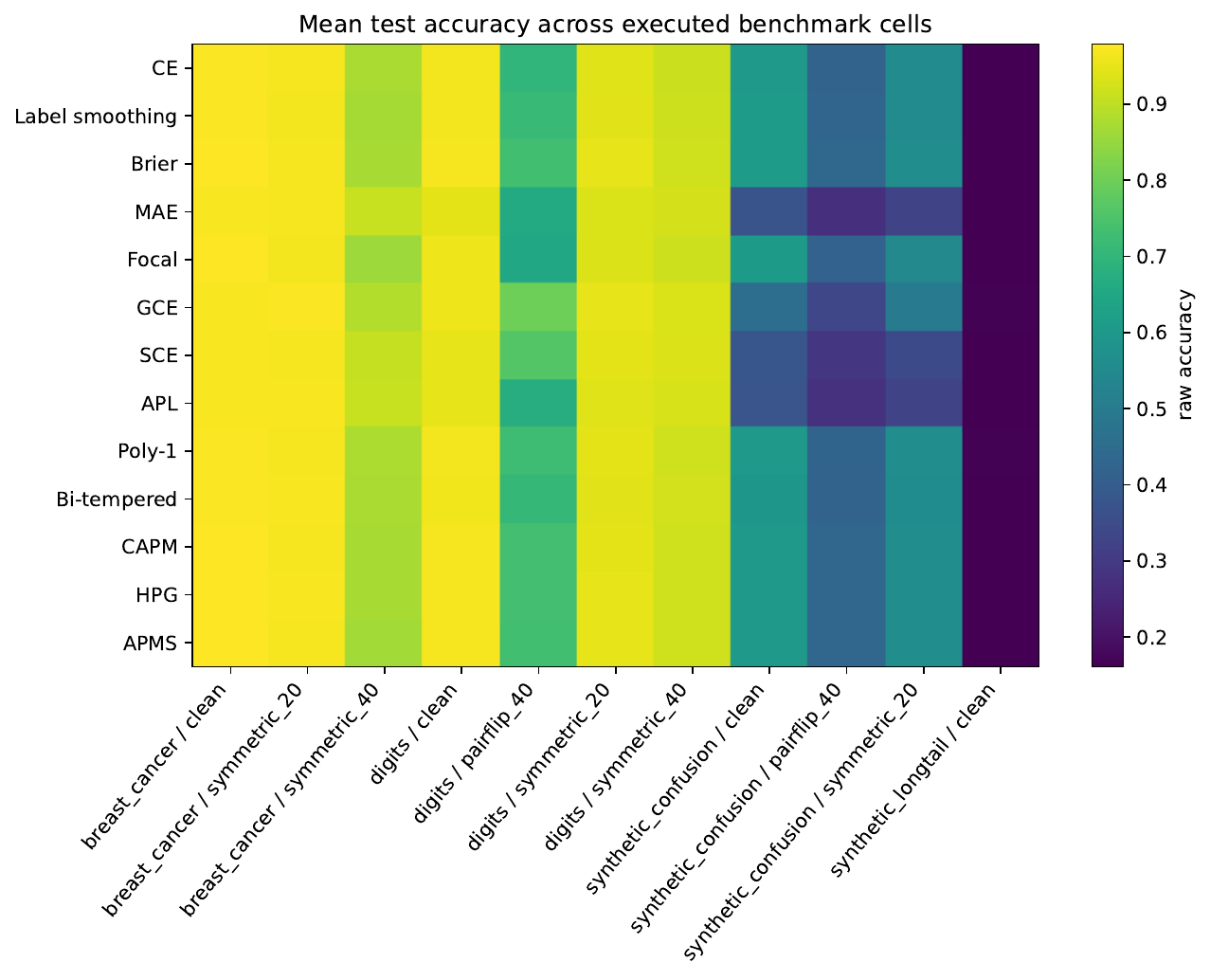}
\end{minipage}\hfill
\begin{minipage}{.32\textwidth}\centering
\includegraphics[width=\linewidth]{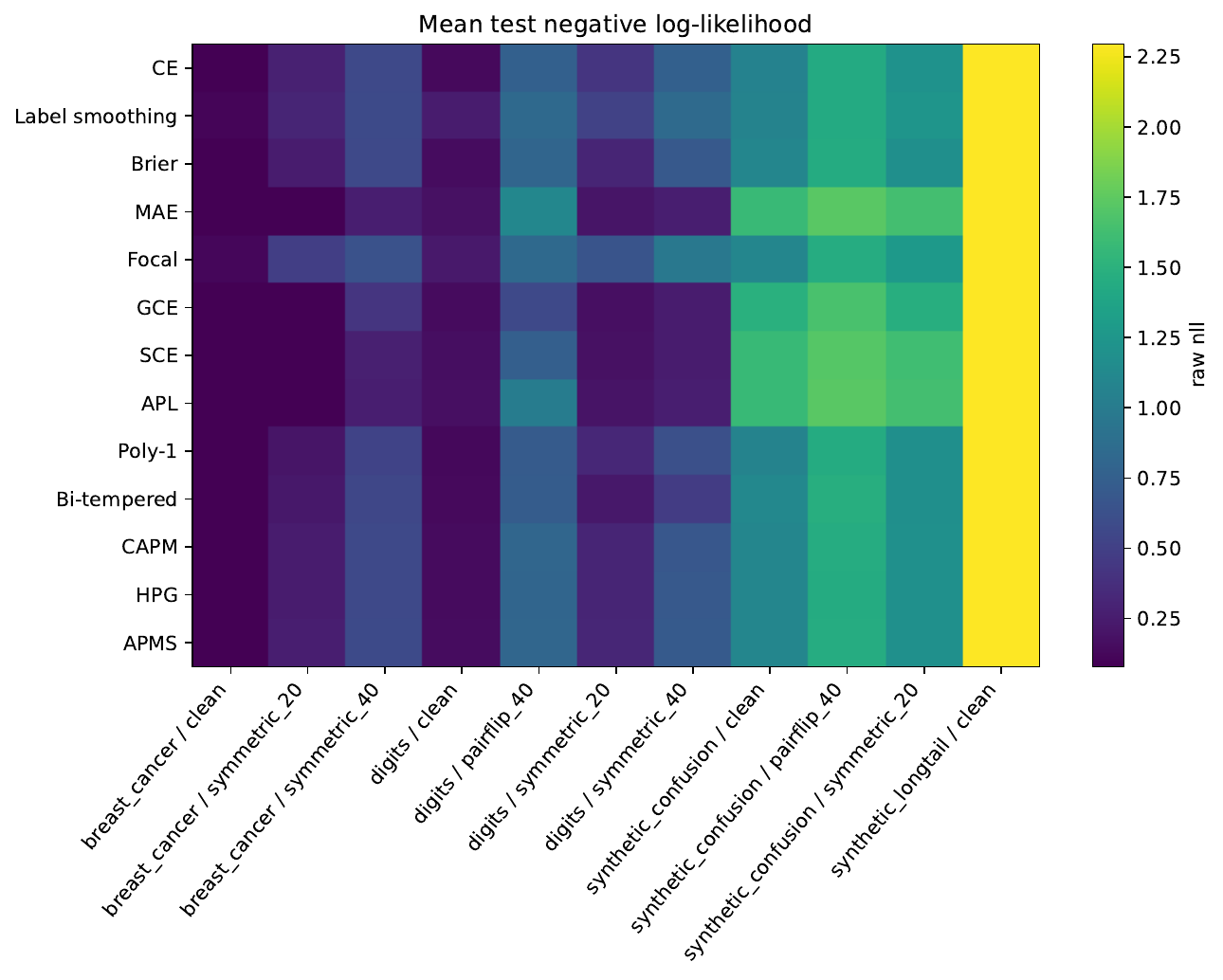}
\end{minipage}\hfill
\begin{minipage}{.32\textwidth}\centering
\includegraphics[width=\linewidth]{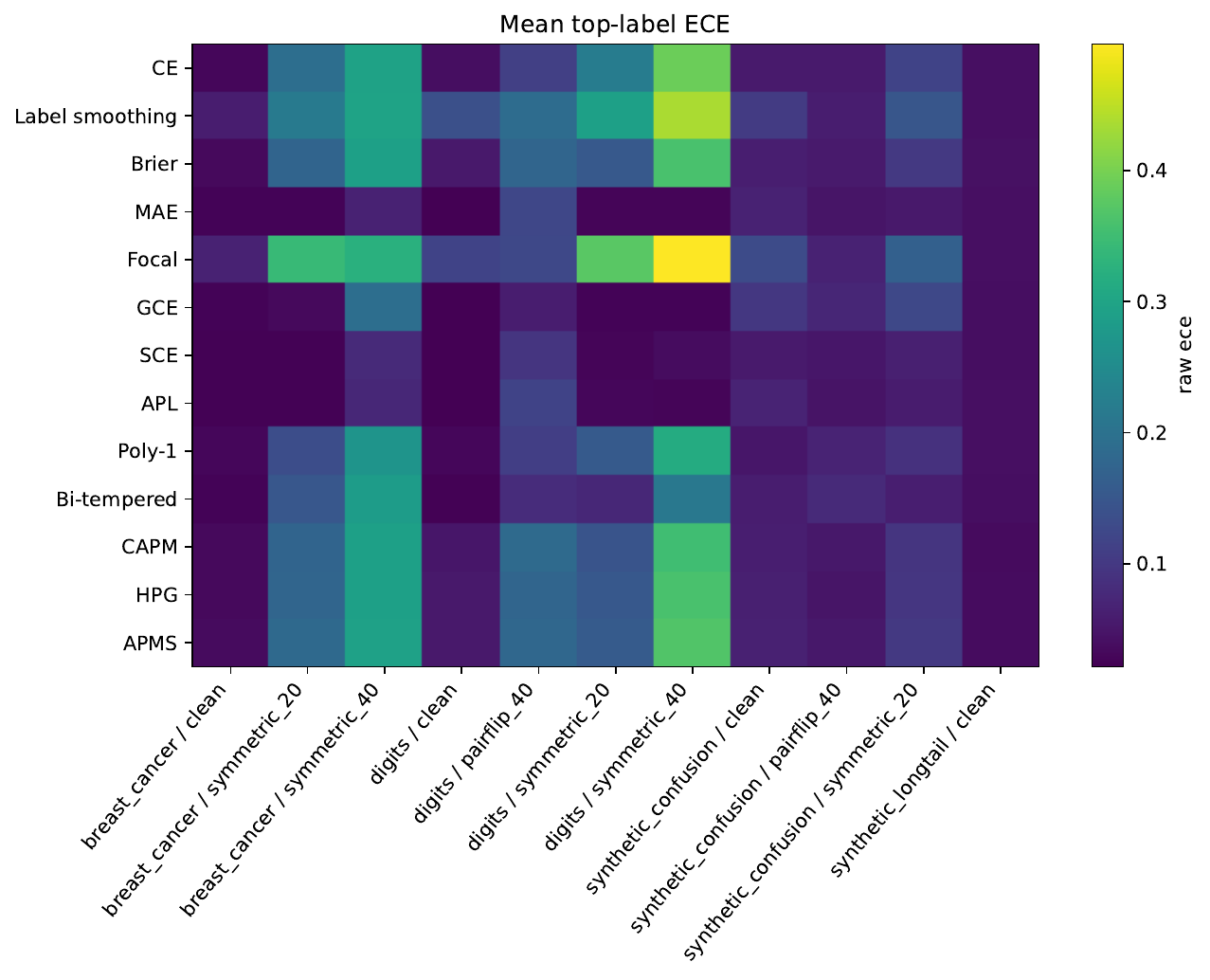}
\end{minipage}
\caption{Aggregate empirical heatmaps. Columns are dataset--training-regime cells, rows are losses, and colors encode retained-run means for accuracy, negative log-likelihood, and top-label ECE, respectively. Accuracy is better when larger; NLL and ECE are better when smaller, although ECE should be interpreted jointly with accuracy and proper-score metrics.}
\label{fig:supp-metric-heatmaps}
\end{figure*}

\begin{figure*}[p]
\centering
\begin{minipage}{.49\textwidth}\centering
\includegraphics[width=\linewidth]{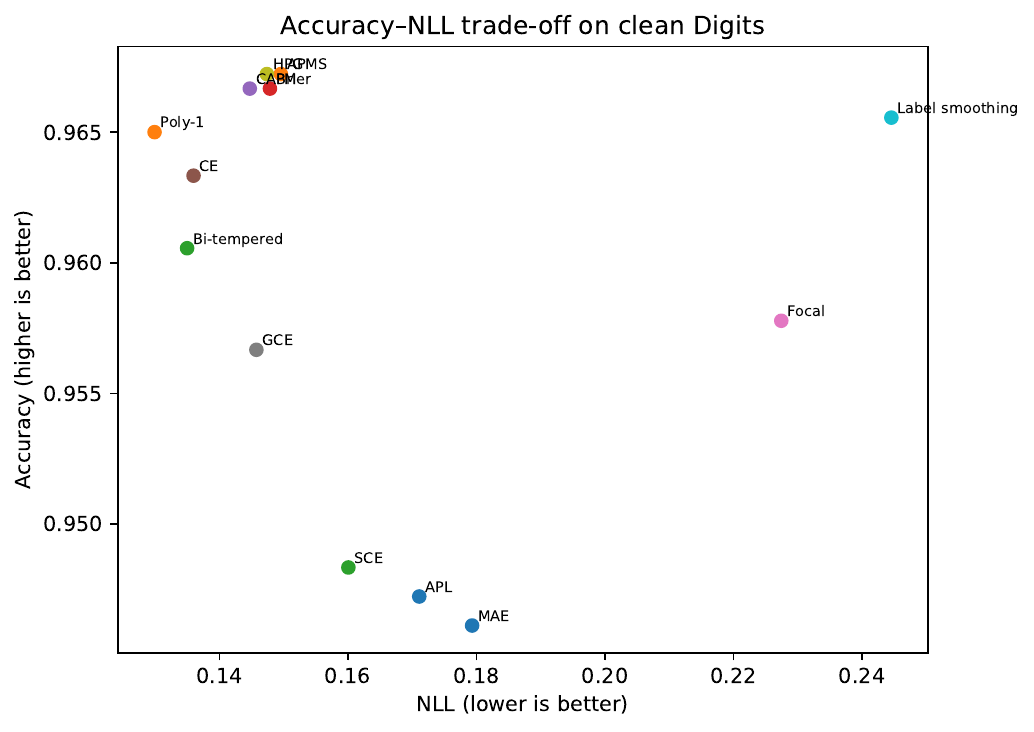}
\end{minipage}\hfill
\begin{minipage}{.49\textwidth}\centering
\includegraphics[width=\linewidth]{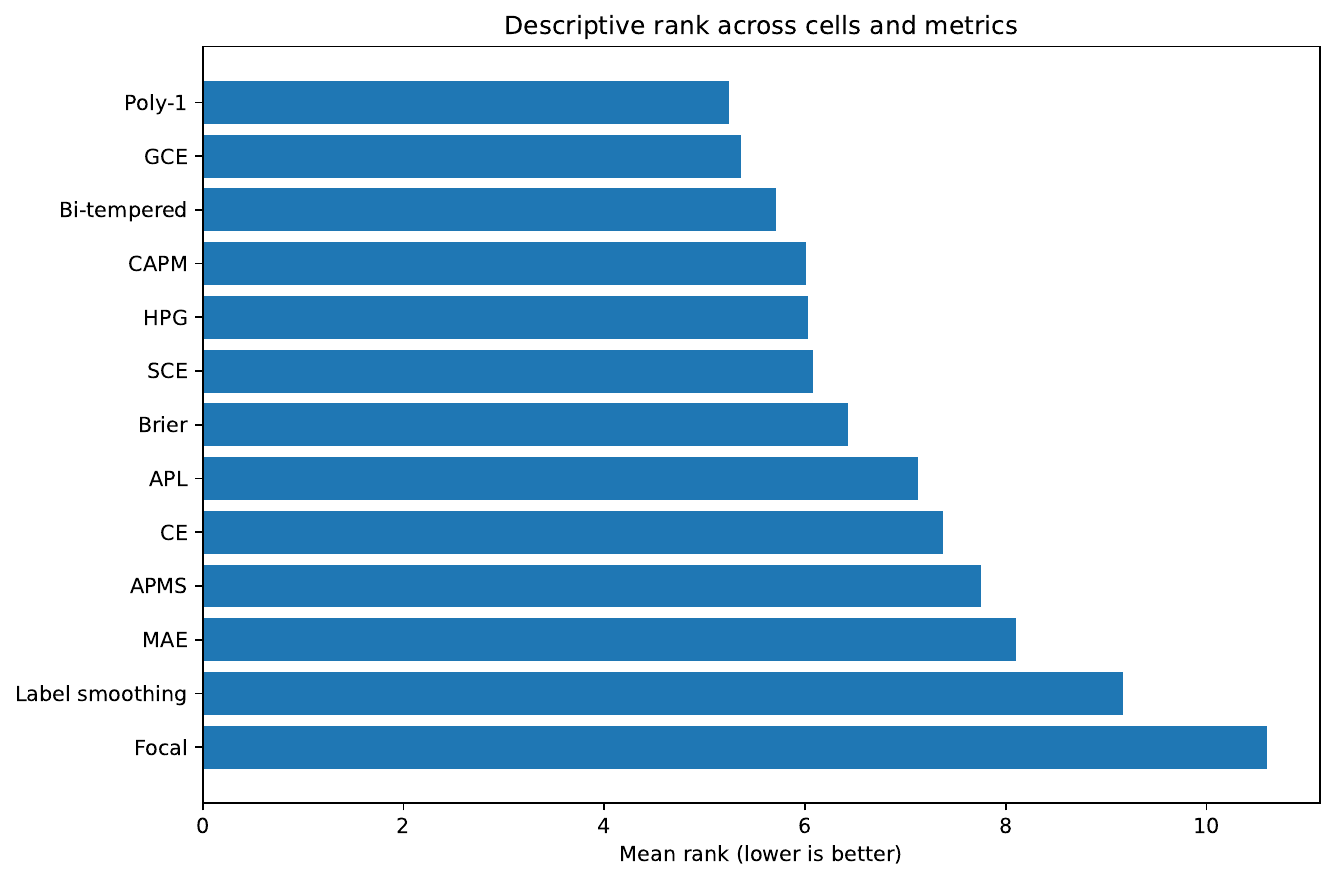}
\end{minipage}
\caption{Clean-Digits and rank summaries. The left panel plots each loss by clean-Digits test NLL on the horizontal axis and test accuracy on the vertical axis, so the upper-left region is preferable for those two metrics. The right panel reports descriptive mean rank across retained cells and metrics, where lower rank is better; it is not a statistical significance analysis.}
\label{fig:supp-clean-ranks}
\end{figure*}

\begin{figure*}[p]
\centering
\begin{minipage}{.49\textwidth}\centering
\includegraphics[width=\linewidth]{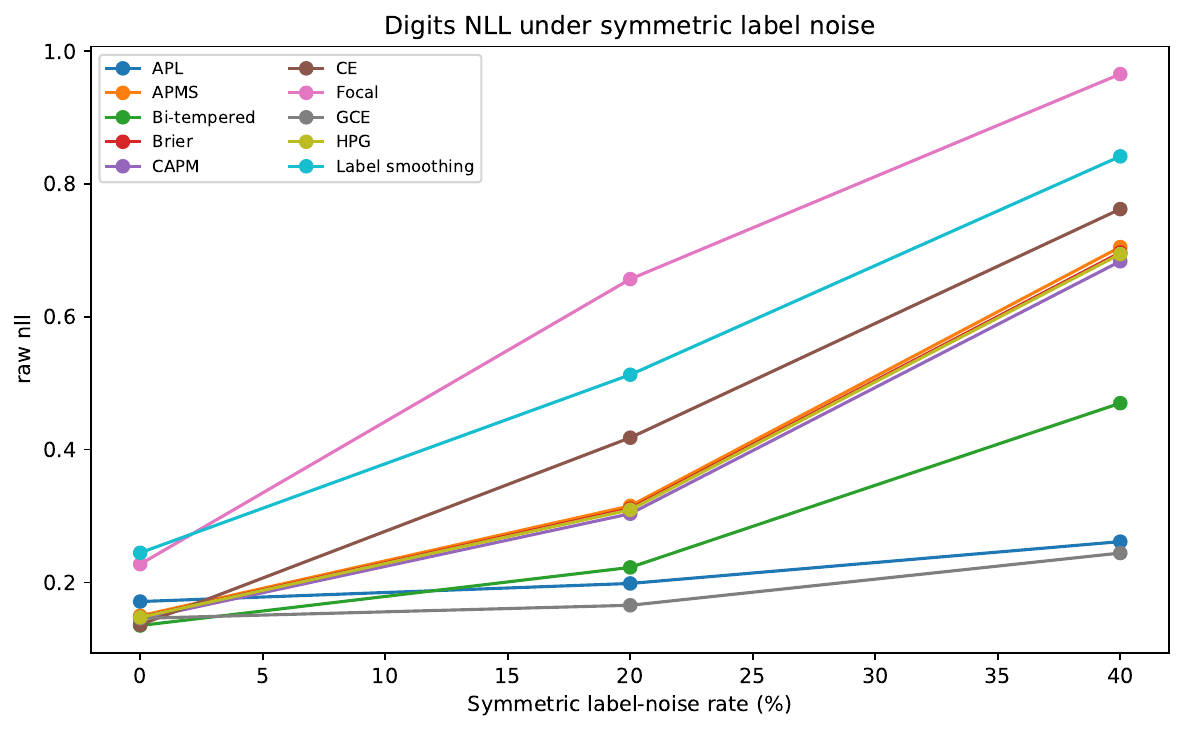}
\end{minipage}\hfill
\begin{minipage}{.49\textwidth}\centering
\includegraphics[width=\linewidth]{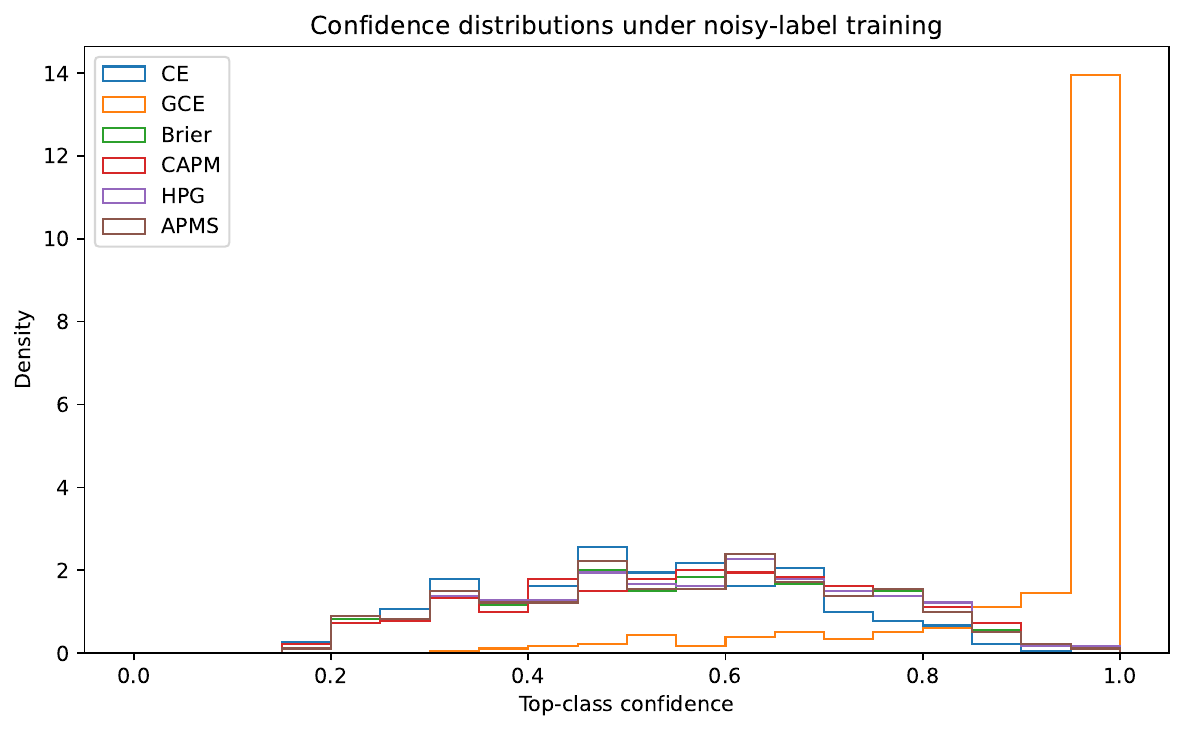}
\end{minipage}
\caption{Additional Digits symmetric-label-noise diagnostics. The left panel shows test NLL as the symmetric training-label corruption rate increases; lower NLL is better. The right panel shows density estimates of top-class confidence after training with 40\% symmetric label corruption for CE, GCE, Brier, CAPM, HPG, and APMS; it describes confidence distributions and does not by itself establish calibration.}
\label{fig:supp-noise-nll-confidence}
\end{figure*}

\begin{figure*}[p]
\centering
\begin{minipage}{.32\textwidth}\centering
\includegraphics[width=\linewidth]{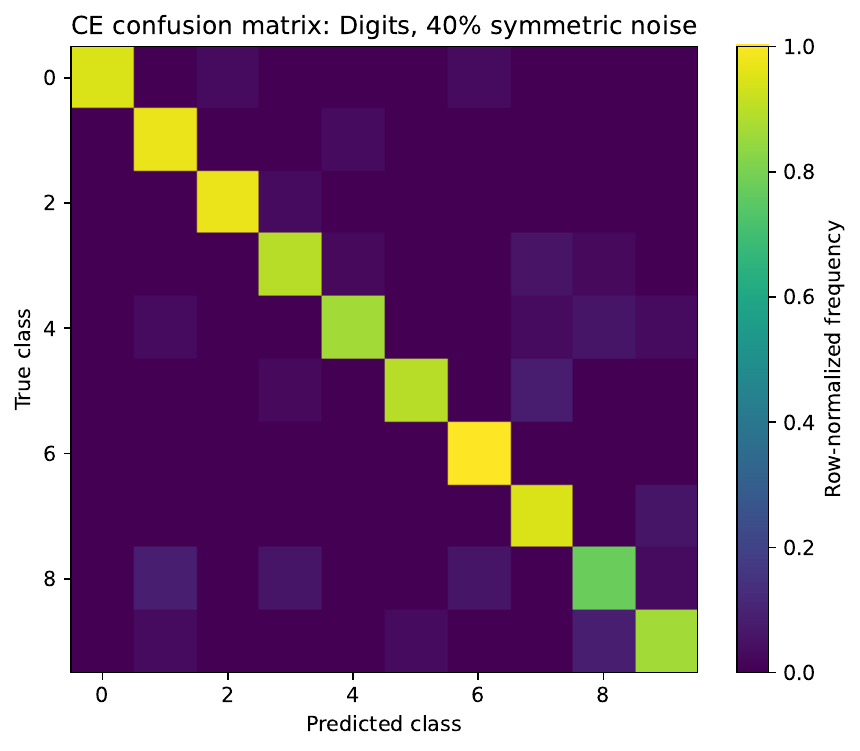}
\end{minipage}\hfill
\begin{minipage}{.32\textwidth}\centering
\includegraphics[width=\linewidth]{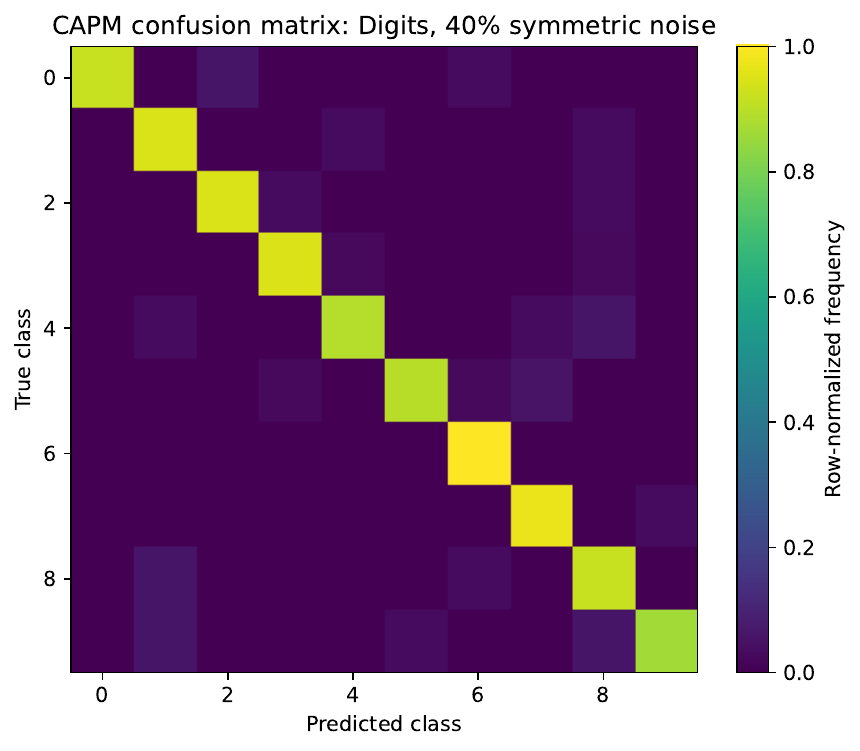}
\end{minipage}\hfill
\begin{minipage}{.32\textwidth}\centering
\includegraphics[width=\linewidth]{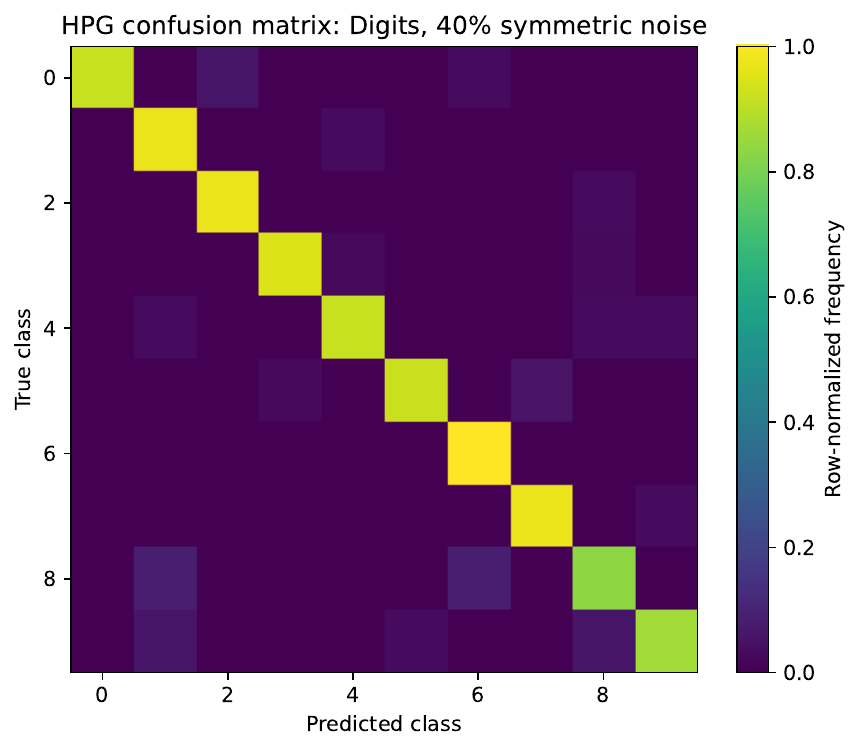}
\end{minipage}

\medskip

\begin{minipage}{.32\textwidth}\centering
\includegraphics[width=\linewidth]{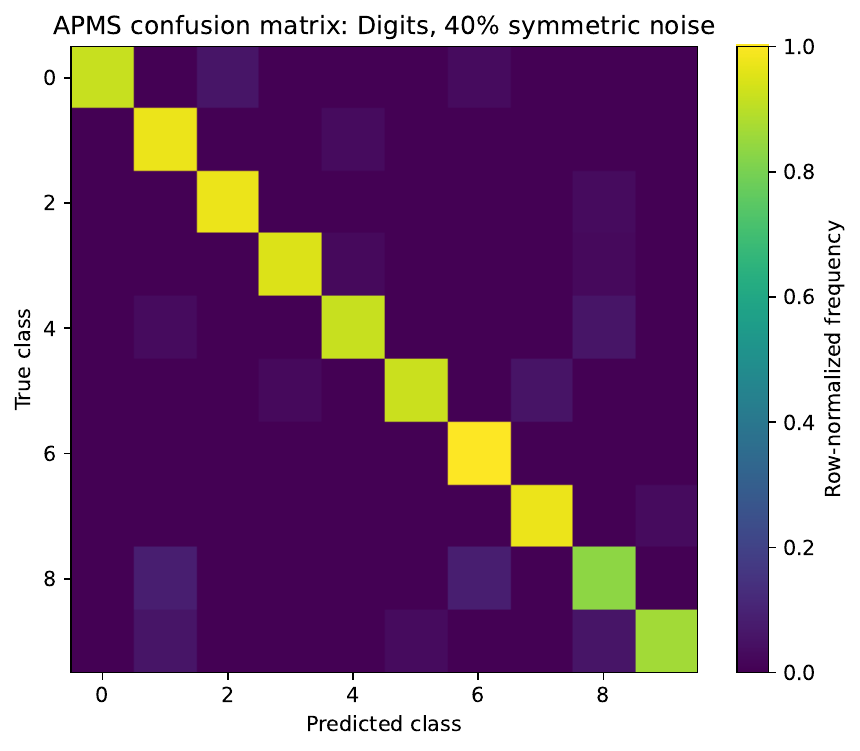}
\end{minipage}\hfill
\begin{minipage}{.32\textwidth}\centering
\includegraphics[width=\linewidth]{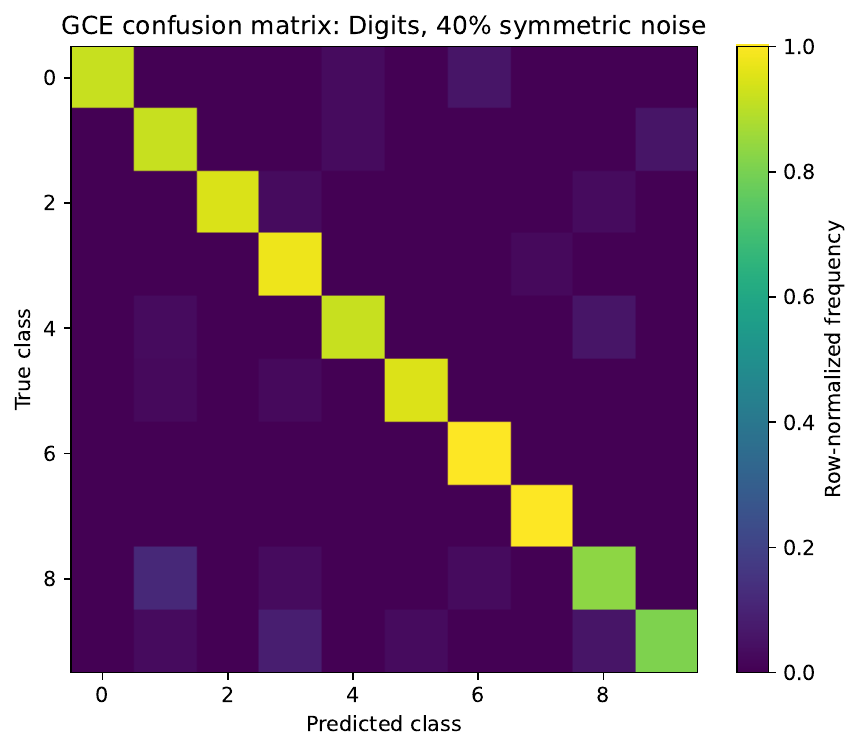}
\end{minipage}
\caption{Digits confusion matrices after 40\% symmetric training-label corruption. Panels show CE, CAPM, HPG, APMS, and GCE, with predicted class on the horizontal axis and true class on the vertical axis. Frequencies are row-normalized, so diagonal mass corresponds to correct predictions within each true class.}
\label{fig:supp-noise-confusions}
\end{figure*}

\begin{figure*}[p]
\centering
\begin{minipage}{.49\textwidth}\centering
\includegraphics[width=\linewidth]{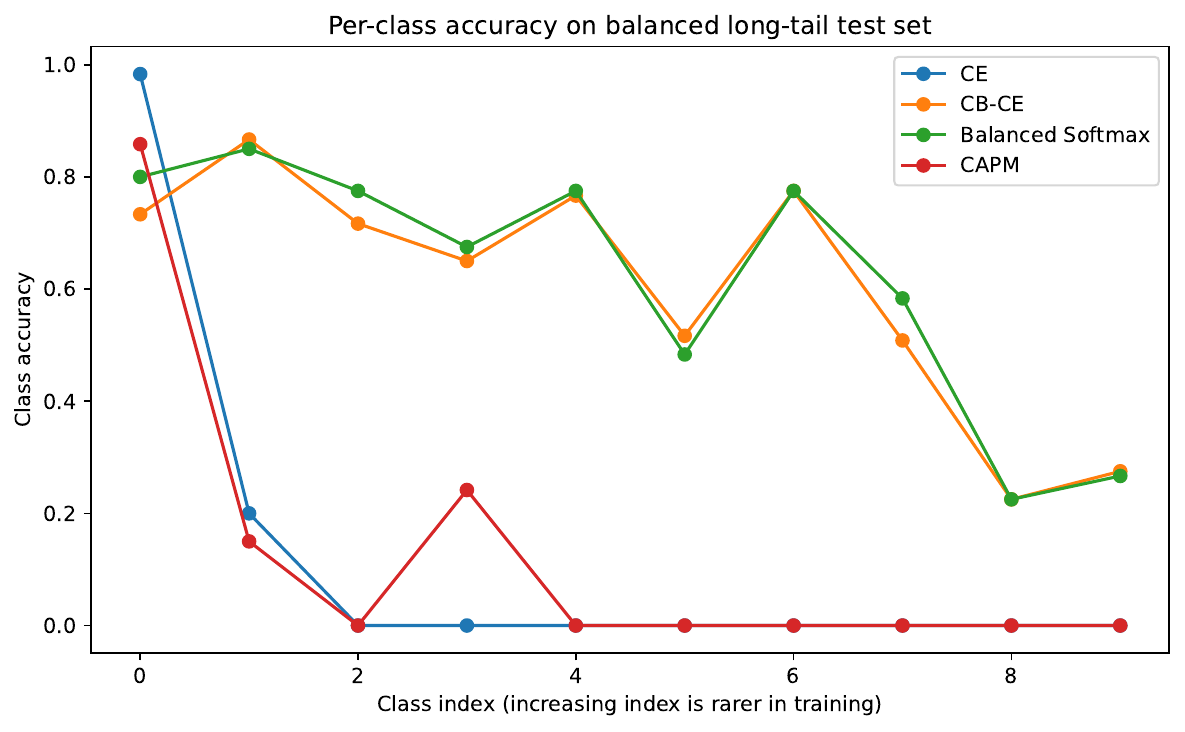}
\end{minipage}\hfill
\begin{minipage}{.49\textwidth}\centering
\includegraphics[width=\linewidth]{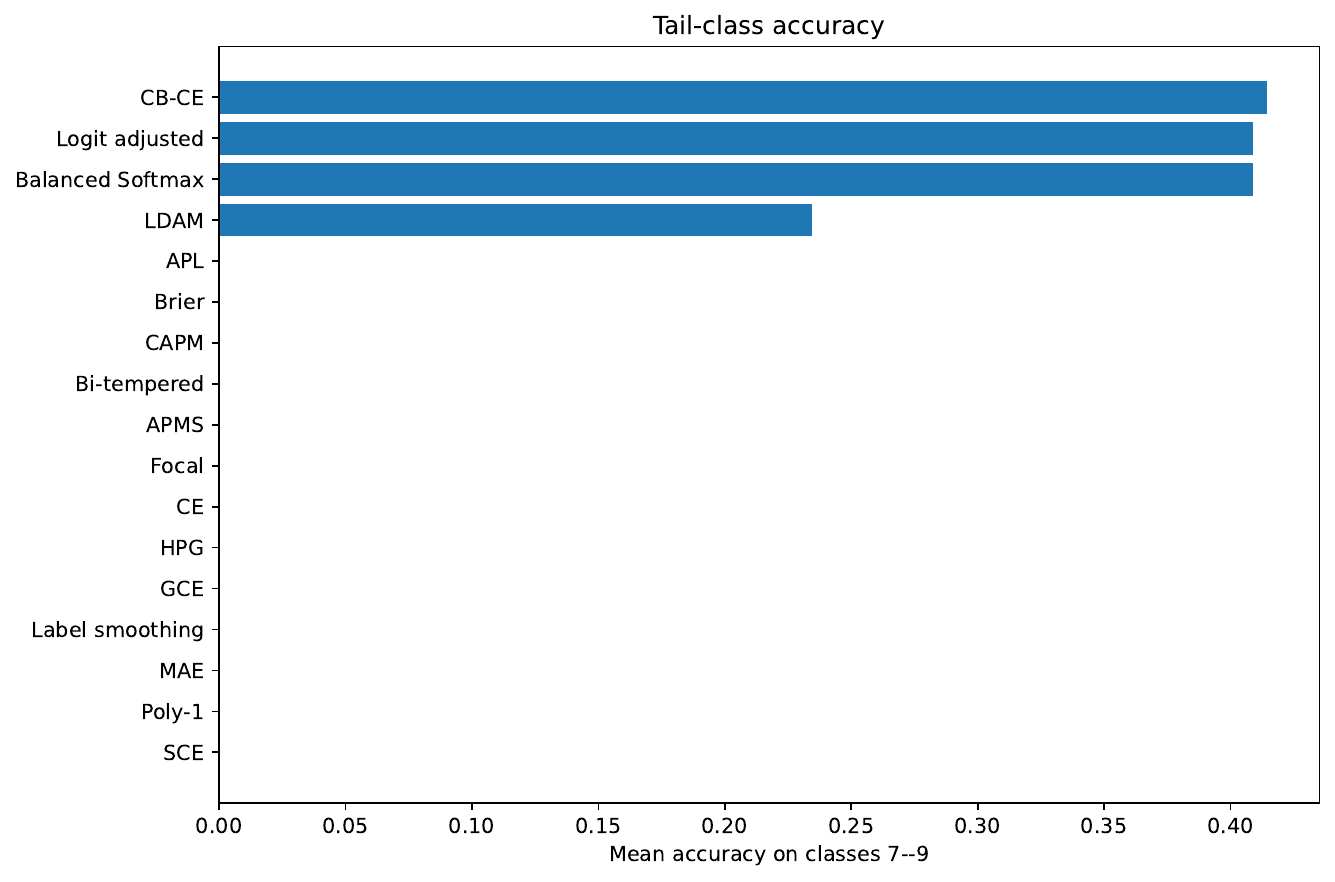}
\end{minipage}
\caption{Long-tail diagnostics on the balanced synthetic test split. The left panel shows per-class accuracy for representative methods, with larger class index indicating rarer classes in the imbalanced training sample. The right panel summarizes mean accuracy on tail classes 7--9 across losses; higher values indicate better tail-class recognition in this synthetic design.}
\label{fig:supp-longtail-details}
\end{figure*}

\begin{figure*}[p]
\centering
\begin{minipage}{.49\textwidth}\centering
\includegraphics[width=\linewidth]{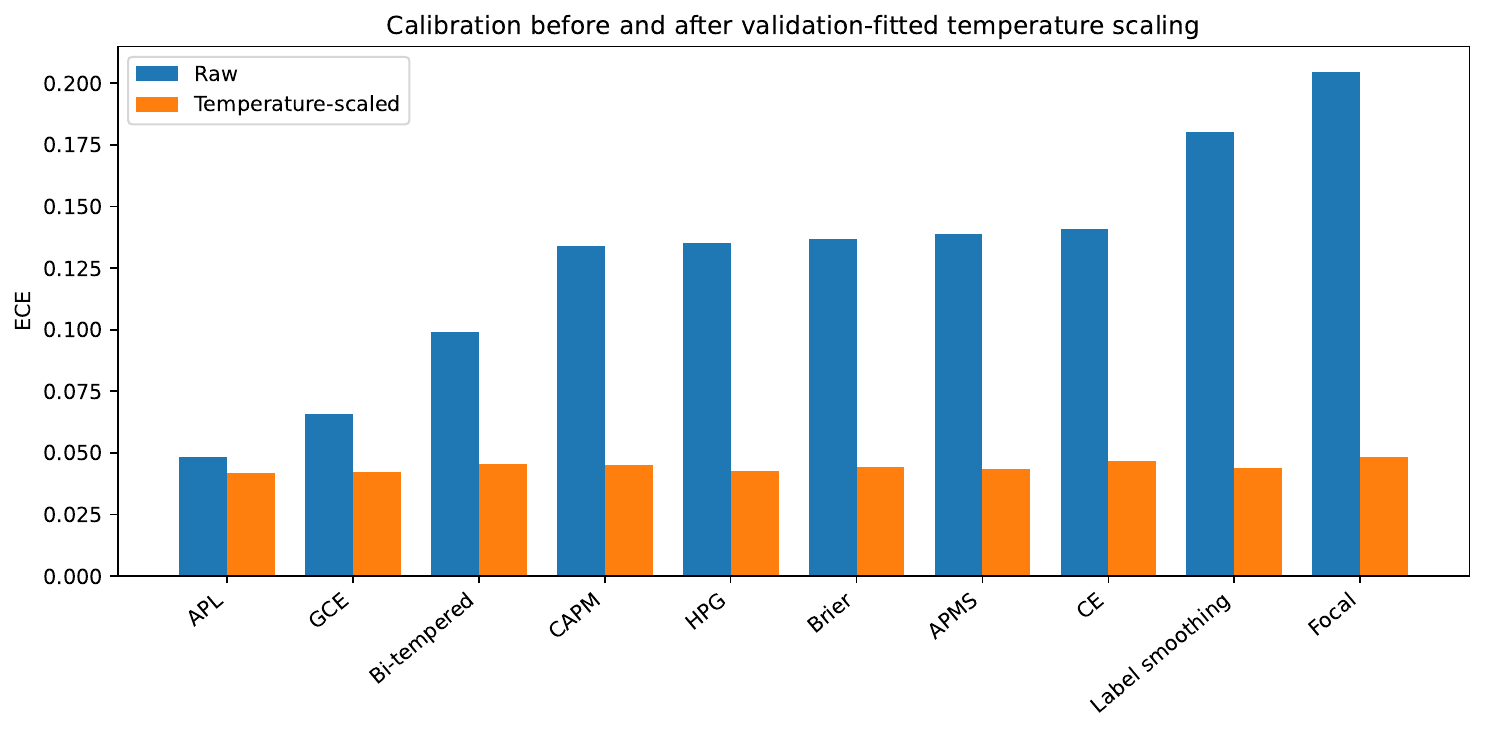}
\end{minipage}\hfill
\begin{minipage}{.49\textwidth}\centering
\includegraphics[width=\linewidth]{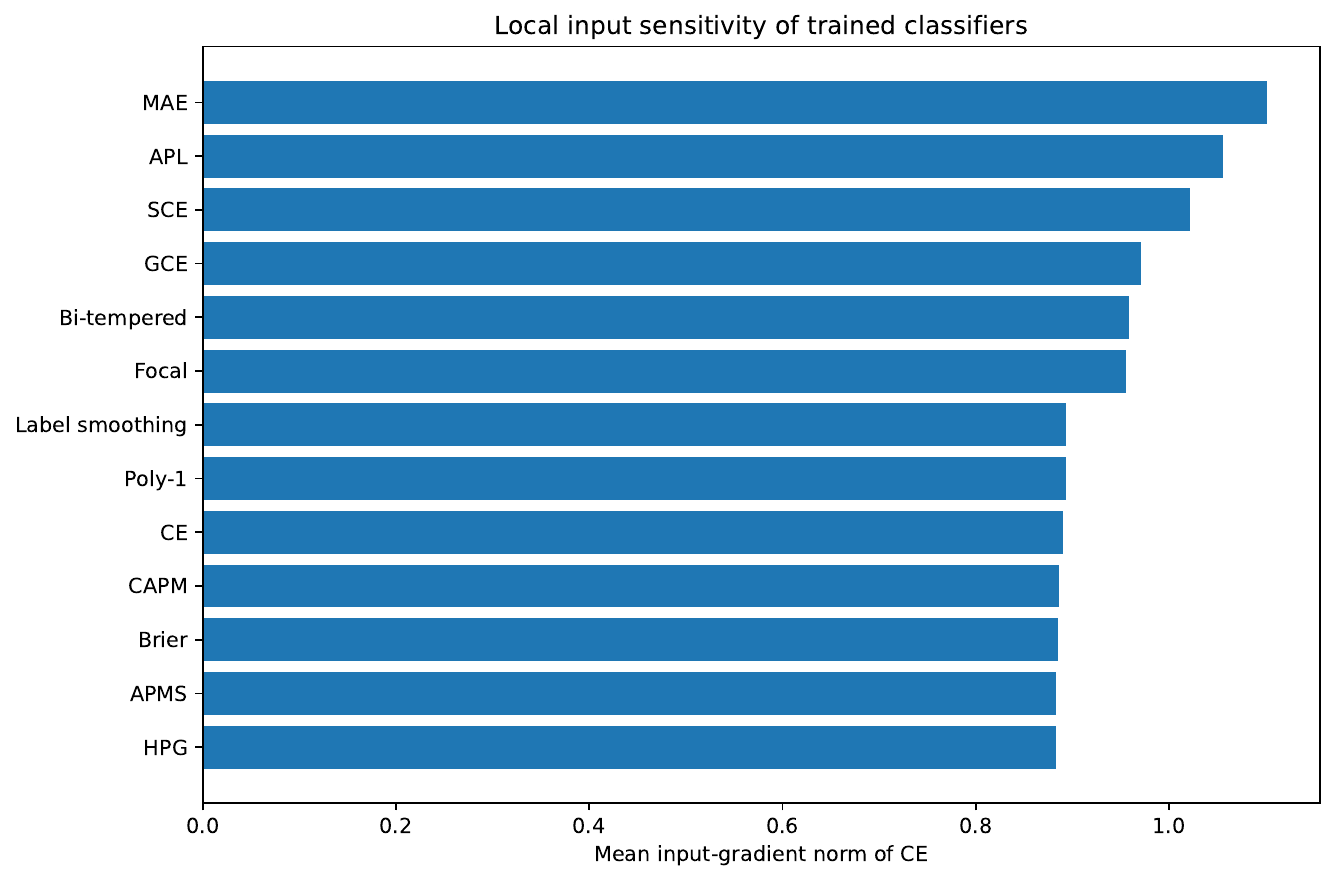}
\end{minipage}
\caption{Calibration and local-sensitivity diagnostics. The left panel compares raw and validation-temperature-scaled ECE for the retained losses; lower ECE is better but should be interpreted with NLL and accuracy. The right panel compares mean input-gradient norms of trained classifiers using the normalization indicated on the horizontal axis; it is a local sensitivity summary, not an adversarial robustness certificate.}
\label{fig:supp-calibration-gradients}
\end{figure*}

\begin{figure*}[p]
\centering
\begin{minipage}{.49\textwidth}\centering
\includegraphics[width=\linewidth]{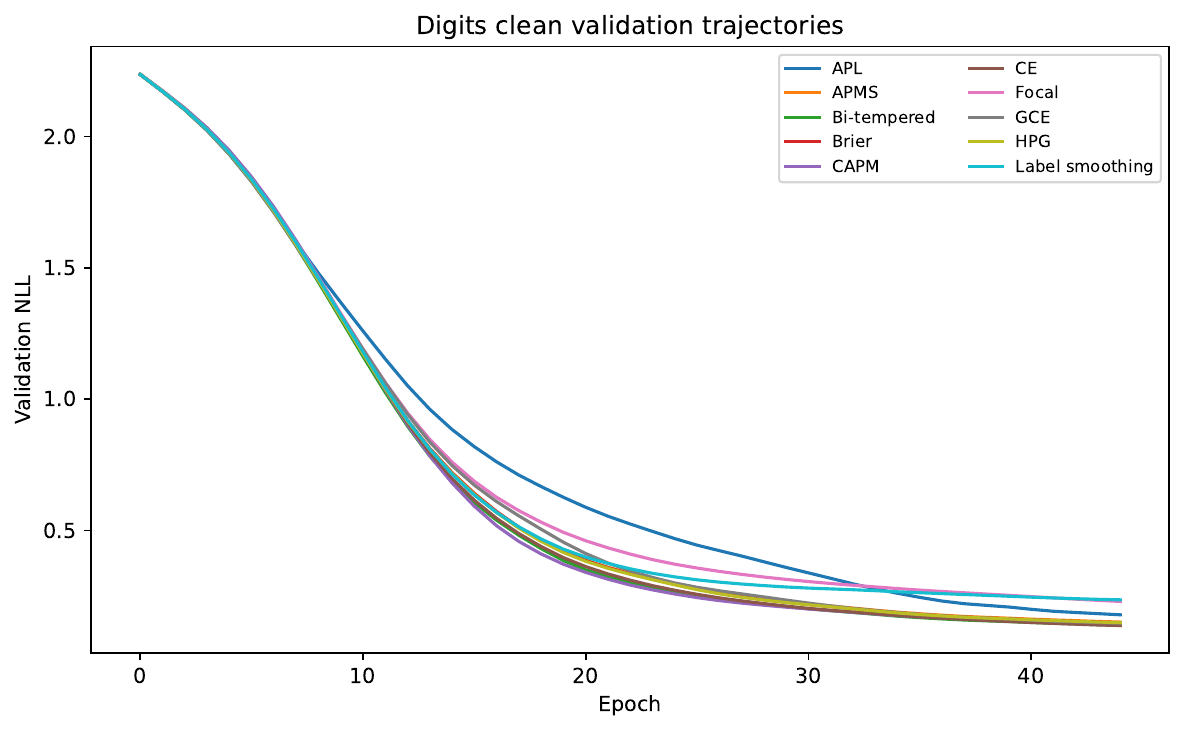}
\end{minipage}\hfill
\begin{minipage}{.49\textwidth}\centering
\includegraphics[width=\linewidth]{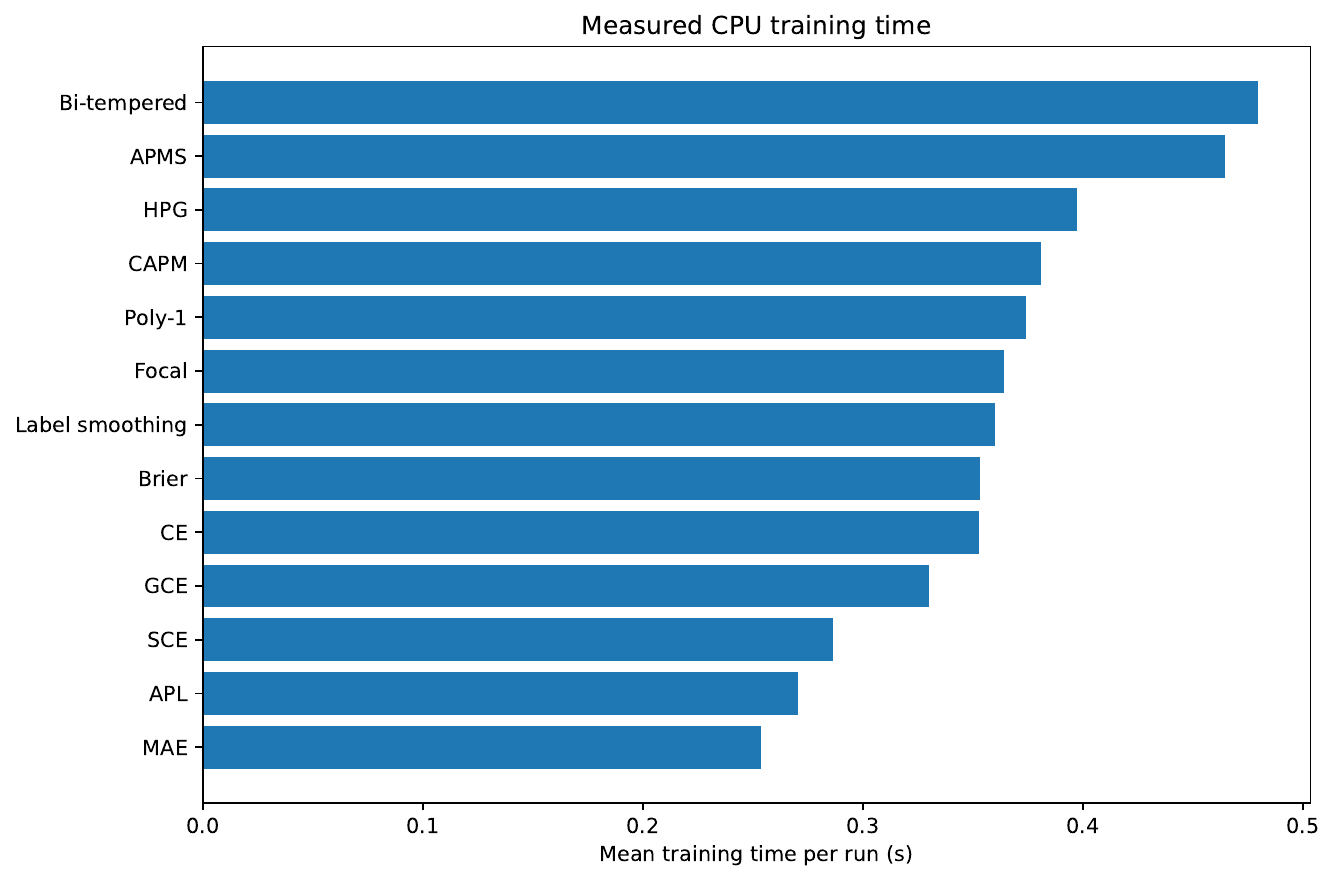}
\end{minipage}
\caption{Optimization diagnostics. The left panel shows clean-Digits validation-NLL trajectories over epochs for the retained losses, with lower validation NLL preferred for checkpoint selection. The right panel reports measured mean CPU training time per run; lower runtime means faster training under the retained implementation and hardware conditions.}
\label{fig:supp-optimization}
\end{figure*}

\clearpage
\end{document}